\documentclass[final,3p,times]{elsarticle}

\usepackage{amssymb}
\usepackage{amsmath}
\newtheorem{definition}{Definition}
\usepackage{subcaption}
\usepackage{setspace}
\journal{Pattern Recognition}

\begin{document}

\begin{frontmatter}



\title{Topological safeguard for evasion attack interpreting the neural networks' behavior}


\author{Xabier Echeberria-Barrio\fnref{label1,label2}}
\ead{xetxeberria@vicomtech.org}
\author[label1]{Amaia Gil-Lerchundi}
\ead{agil@vicomtech.org}
\author[label2]{Iñigo Mendialdua}
\ead{inigo.mendialdua@ehu.eus}
\author[label1]{Raul Orduna-Urrutia}
\ead{rorduna@vicomtech.org}
\affiliation[label1]{organization={Vicomtech Foundation, Basque Research and Technology Alliance (BRTA)},addressline={Mikeletegi 57},city={Donostia-San Sebastian},postcode={20009},country={Spain}}

\affiliation[label2]{organization={Department of Computer Languages and Systems, University of the Basque Country (UPV/EHU)},city={Donostia-San Sebastian},postcode={20018},country={Spain}}

\begin{abstract}
In the last years, Deep Learning technology has been proposed in different fields, bringing many advances in each of them, but identifying new threats in these solutions regarding cybersecurity. Those implemented models have brought several vulnerabilities associated with Deep Learning technology. Moreover, those allow taking advantage of the implemented model, obtaining private information, and even modifying the model's decision-making. Therefore, interest in studying those vulnerabilities/attacks and designing defenses to avoid or fight them is gaining prominence among researchers. 
In particular, the widely known evasion attack is being analyzed by researchers; thus, several defenses to avoid such a threat can be found in the literature. Since the presentation of the L-BFG algorithm, this threat concerns the research community. However, it continues developing new and ingenious countermeasures since there is no perfect defense for all the known evasion algorithms. 
In this work, a novel detector of evasion attacks is developed. It focuses on the information of the activations of the neurons given by the model when an input sample is injected. Moreover, it puts attention to the topology of the targeted deep learning model to analyze the activations according to which neurons are connecting. This approach has been decided because the literature shows that the targeted model's topology contains essential information about if the evasion attack occurs. For this purpose, a huge data preprocessing is required to introduce all this information in the detector, which uses the Graph Convolutional Neural Network (GCN) technology. Thus, it understands the topology of the target model, obtaining promising results and improving the outcomes presented in the literature related to similar defenses.
\end{abstract}



\begin{keyword}
Artificial neural network interpretability\sep Artificial neural network cybersecurity\sep Adversarial learning\sep Evasion attack\sep Artificial neural network countermeasure 



\end{keyword}

\end{frontmatter}


\section{Introduction}
\label{sec:introduction}

The improvements in the computing power capacities have meant that deep neural networks are being extensively studied and implemented in many fields that directly impact humans' lives. Even those technologies are being introduced in critical areas since any decision-making mistake could potentially compromise people's lives. Healthcare \cite{finlayson2019adversarial} and autonomous vehicles \cite{Sharma2019} are examples where any error can be fateful. That concern leads researchers to study and analyze possible defenses on deep neural network models discovering countermeasures against known threats. The literature has shown several vulnerabilities that a deep learning model can suffer and how it can be corrupted or manipulated \cite{9252914}.  

There exist a worrisome threat called evasion attack \cite{jiang2020poisoning}. It has been widely studied, and attempts have been made to develop different defenses to make models more robust against it. In this attack, the attacker takes advantage of the target model's sensitivity by adding a specifically designed noise to an input sample. Indeed, it can modify the original output prediction of the sample, even though this distortion is imperceptible for humans. The first algorithm generating an adversarial example in a deep neural network was L-BFGS  \cite{Szegedy2014IntriguingNetworks}. It was followed by the development of more efficient and less detectable algorithms. Even some of the developed algorithms allow for generating adversarial attacks that are transferable from other models \cite{Dong2018}. Among those improved evasion algorithms are distinguished the followings: Fast Gradient Sing Method (FGSM) \cite{Madry2018TowardsAttacks}, Basic Iterative Method (BIM) \cite{Kurakin2019AdversarialWorld}, Projected Gradient Descent Method (PGD) \cite{Madry2018TowardsAttacks}, and Carlini-Wagner Method \cite{7958570}. To defend the neural networks from this threat, the researchers have proposed several countermeasures, making them more robust \cite{Echeberria-Barrio2021} or adding external detectors \cite{PAWLICKI2020148}.

Taking the investigation of Echeberria-Barrio et al. \cite{echeberria2022understanding} and studying its results, they showed that in the presented graphs, there is information about the behavior of a deep learning model that can help in the detection of adversarial examples. In other words, it can help to develop a successful defense against evasion attacks in deep learning models. The presented graph type represents the classifier part of a deep learning model, i.e., a fully connected multi-layer perceptron (MLP). Those graphs take the neurons as nodes, and the edges are the connections between those neurons. Moreover, the edges are weighted by the source neuron's activation values. It is already known that the activation values can give information about if the input data is adversarial or not, making some of those detectable \cite{PAWLICKI2020148}. However, Echeberria-Barrio et al. \cite{echeberria2022understanding} introduce a new neural attribute (different from activation value) called the impact of the neuron. It is computed via the input and output activation values of the neuron. One of the differences between the activation value and the impact of a neuron is that the second one contains quit more topological information about the neural network. Moreover, Echeberria-Barrio et al. \cite{echeberria2022understanding} show how the impact attribute contains information about if input data is adversarial and highlights the topology of the classifier (the neighbourhood of a neuron) as information against evasion attack.

\begin{figure}[!htp]
    \centering
    \includegraphics[width=0.46\textwidth]{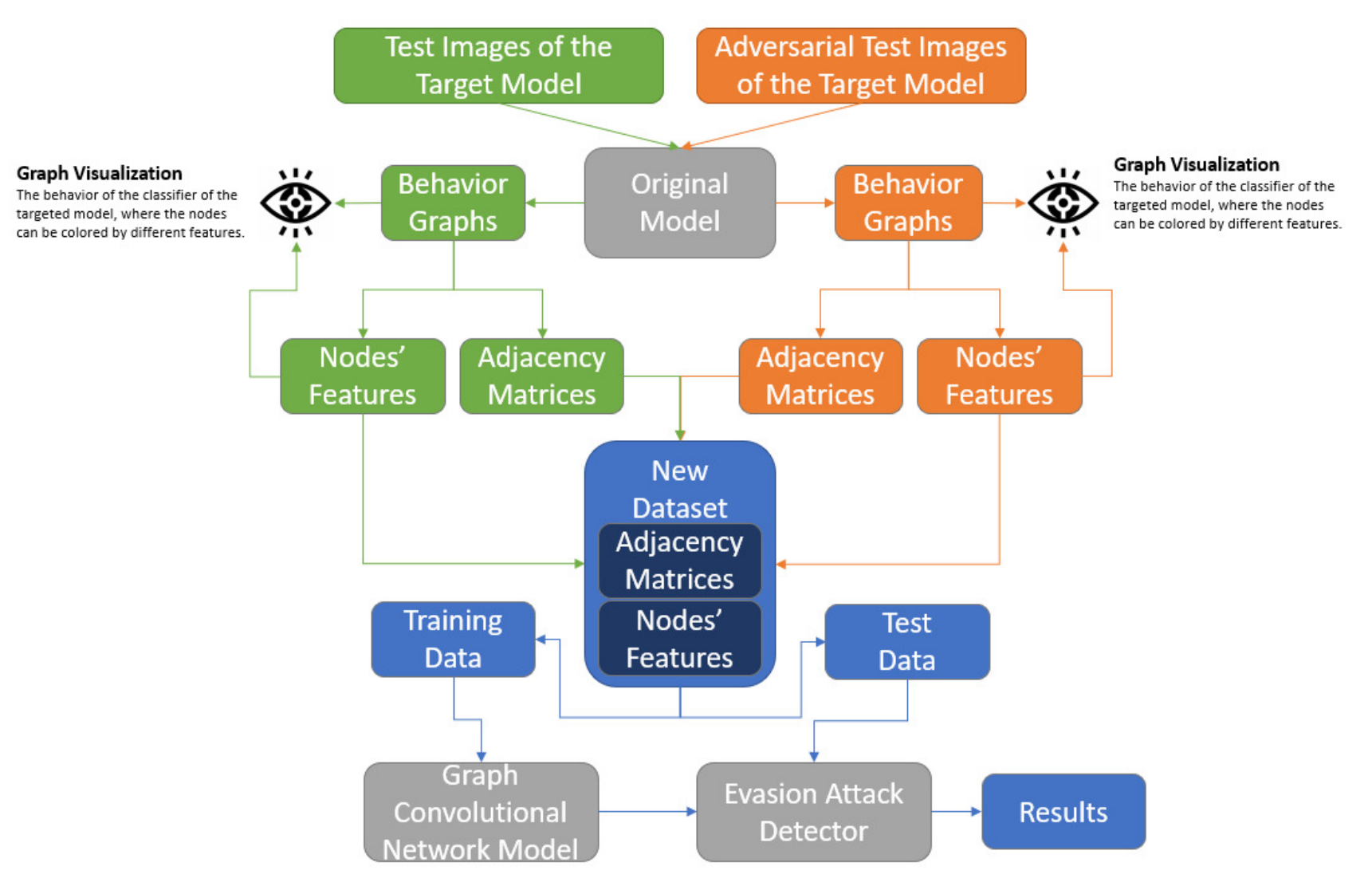}
    \caption{Diagram of the processes that are followed in this work to develop the detector against the evasion attack.}
    \label{fig:conexions}
\end{figure}

This work presents a novel evasion attack detector based on graph neural network technology. This detector focuses on the classifier of a target deep learning model, and it requires a preprocessing of this model's input data to identify if it is adversarial or not. The preprocessing is based on the behavior graph definition presented in \cite{echeberria2022understanding}. Moreover, the importance of the topological information in a deep learning model is given the decision to use graph neural network technology. This work aims to develop an evasion attack detector, improving some detection rates found in the literature and presenting a new way to develop novel detectors in this field. In addition, this type of detector can provide detailed information about the neurons of the target deep learning model, but it will be studied in future works. This work starts by introducing the novel evasion attack detector and all the required technical notions. Then, it tests the developed system in a specific scenario, obtaining performance rates. Moreover,  this work continues evaluating the mentioned topological information to analyze the information it gives to the presented detector. Finally, the obtained performance rates are compared with qualified detectors found in the literature \cite{aldahdooh2022adversarial} and the novel evasion detector is theoretically compared with a similar detector found in the literature introduced by Pawlicki et al.\cite{PAWLICKI2020148}.   

The rest of the paper is divided as follows: Section \ref{sec:literature} details related literature, concretely the evasion attack detection methods applied to deep learning models. Section \ref{sec:methodology} explains the classifier behavior graph of the fully connected part of a deep learning model and defines all the necessary attributes of the nodes. Section \ref{sec:experiment} describes how the evasion attack detector is developed and how the essential dataset is generated. Finally, the results and discussion are given in Section \ref{sec:results} and Section \ref{sec:conclusion} lists the learned lessons and future work.

\section{Literature Review}\label{sec:literature}

Several possible defenses against the evasion attack have been put forward. On the one hand, one widely known defense is adversarial training, which tries to correctly classify the corrupted input itself or create a separate class indicating that the input sample is manipulated. For example, Yu et al. \cite{YU2023109054} developed a sophisticated adversarial training method that generates a particular noised dataset according to the target model to retrain it. Moreover, recently Pintor et al. \cite{PINTOR2023109064} presented a variation of this famous defense, which consists of the generation of a generic dataset based on the Imagenet dataset that implements the adversarial training in the target model since the first training. On the other hand, defenses exist that introduce a dimensional reduction into the original model to avoid the extra noise injected by the adversary. For example, Bhagoji et al. \cite{bhagoji2018enhancing} discuss the incorporation of the  Principal Component Analysis algorithm to generate more robust models, while Sahay et al. \cite{ sahay2019combatting} introduce the autoencoder neural network architecture to obtain the desired dimensional reduction in the input data, defending the model from adversarial examples. Those methods have improved vulnerable models, making them more robust. However, they usually cause side effects on the model in many applications, reducing the targeted model's accuracy or modifying it.

A different way of countering the problem comes as a detector without modifying the targeted model's behavior and architecture. Several approaches propose a second classifier to distinguish adversarial examples from non-adversarial ones using the same distribution \cite{ma2019nic}. For example, Feinman et al. \cite{feinman2017detecting} investigate the Bayesian uncertainty estimates of the confidence on the adversarial samples. Otherwise, Ma et al. \cite{ma2018characterizing} present the LID algorithm, which assesses if the input sample is malicious, analyzing its distance distribution to its neighbors. Two years later, Kherchouche et al. \cite{kherchouche2020detection} introduce the NSS algorithm, which extracts particular statistic properties from the targeted input sample that are altered by the presence of adversarial perturbations. Even though these types of detectors have good results, they can be evaded by formulating the attack to find adversarial examples that fool both classifiers simultaneously, as demonstrated by Carlini et al. \cite{10.1145/3128572.3140444}.

Other types of detectors have been developed where they work with different information of the model and input samples, avoiding the problem mentioned above. Similar approaches to the one suggested in this paper were presented by Pawlicki et al. \cite{PAWLICKI2020148} and by Mertzen et al. \cite{metzen2017detecting}. On the one hand, Pawlicki et al. \cite{PAWLICKI2020148} develop an evasion attack detector focused on analyzing all the activations generated in the target model. On the other hand, Mertzen et al. \cite{metzen2017detecting} generate new branches in the main network at some layer in the ResNet \cite{he2016deep} Convolutional Neural Network, which produce a probability of the input being adversarial. It performs well in Cifar10 \cite{krizhevsky2009learning} and Imagenet \cite{russakovsky2015imagenet} datasets.

Finally, no references have been found that reduce the number of activations of the targeted model to be analyzed for the evasion attack detection. For example, the developed detector may focus on specific neurons of the targeted deep learning model, reducing the activation values to be studied. In addition, no papers have been found that consider those activations, maintaining their topological relations. Echeberria-Barrio et al. \cite{echeberria2022understanding} demonstrate that the topological relations contain information about if a sample is corrupted (evasion attack). However, no detector of evasion attack has been found in the literature that considers the topological relation of the activations of the targeted model to analyze if the input data is corrupted. 

\section{Methodology}\label{sec:methodology}

Suppose $X$ and $Y$ are sets of images and labels, respectively, where for each image $x_{i}\in X$, its corresponding class label is $y_{i}\in Y$. Assume a target deep learning model $\phi$ that classifies $X$ correctly, that is, 
\begin{align*}
  \phi \colon X &\to Y\\
  x_{i} &\mapsto y_{i}
\end{align*}
 Moreover, the targeted model consists of a features-extractor block ($\phi_{F}$), such as a convolutional neural network, a recurrence neural network or an autoencoder, and a dense neural network block ($\phi_{C}$), where the $\phi_{F}$ computes some attributes from the $X$ images, $F^{X}_{\phi} = \{\overrightarrow{f}^{i} = (f^{i}_{1},....,f^{i}_{n})\ |\ \overrightarrow{f}^{i} = \phi_{F}(x_{i}) \}$, and the $\phi_{C}$ classifies $F_{X}$ into $Y$. Therefore, the classifier
\begin{align*}
  \phi_{C} \colon F^{X}_{\phi} &\to Y\\
  \overrightarrow{f}^{i} &\mapsto y_{i}
\end{align*}
decides the class of the images according to the extracted features. Moreover, the layer $l$ of the classifier $\phi_{C}$ will be indicated as $\phi^{l}_{C}$, where it contains the neurons of the layer $l$. Note that the notation presented in this introduction is used during the whole work.

This work presents an adversarial example detector focusing on the analysis of the classifier of the $\phi$, that is $\phi_{C}$. This detector decides if an input image $x_{i}$ is adversarial or not through some attributes computed from the activations of the $\phi_{C}$ that $\overrightarrow{f}^{i}$ generates. This section introduces the novel-developed adversarial examples detector, describing each step needed to obtain the defending system against evasion attack on the model. It begins by introducing the behavior graph for $\phi_{C}$ and giving an example as visualization. Next, several neuron attributes of the $\phi$ are defined, which allows generating a preprocessed enriched new dataset as baseline for the desired evasion attack detector. Once the attributes are presented, the preprocessing method is described. Finally, the detector's technology and architecture are detailed.    

\subsection{Behavior Graph}

\begin{figure}[!ht]
    \centering
    \includegraphics[width=0.46\textwidth]{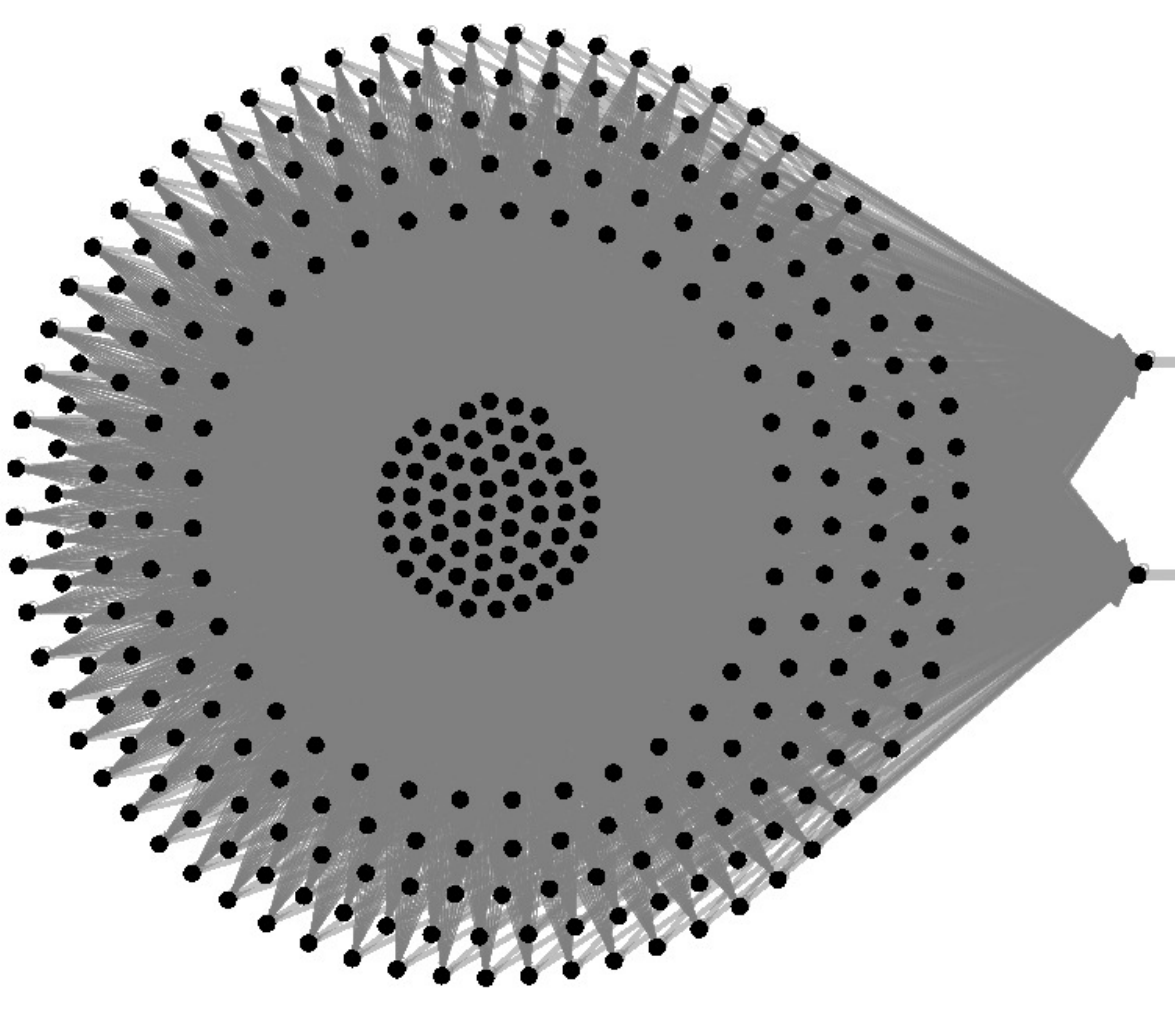}
    \caption{A classifier's behavior graph example, where the central cluster corresponds to the input layer of the classifier, the surrounding nodes are the neurons of the hidden layer, and the right last nodes correspond to the output layer. Notice that the neurons without any connection do not appear. For example, even though the input neuron contains more neurons, they are not visualized in the graph because their activations are zero.}
    \label{fig:example_vis}
\end{figure}

Taking an image $x_{i}\in X$, computing its features, $\phi_{F}(x_{i}) = \overrightarrow{f}^{i}\in F^{X}_{\phi}$, and injecting this vector to classifier, $\phi_{C}(\overrightarrow{f}^{i}) = y_{i}$, a set of activations of $\phi_{C}$ is obtained, noted by $A^{x{i}}_{\phi_{C}}$. Those activations connect two neurons in  $\phi_{C}$, and they can be understood as the weights of the edges between the neurons, whereas the neurons can be considered nodes. Therefore, those activations and the neurons of the classifier give a graph associated with an image $x_{i}\in X$ of the classifier, $\phi_{C}$, of the targeted model (Figure \ref{fig:example_vis}). This graph shows how the neurons of $\phi_{C}$ are activating and where those have more or less influence. Those activation values and topological relations are the reason for the decision-making of the targeted deep learning model. In other words, the information in the presented graph determines the behavior of the targeted model. That is why this graph is named \textit{behavior graph}.    

\begin{definition}[Behavior Graph]
Let $\phi$ be a deep learning model containing $\phi_{F}$ feature-extractor block and $\phi_{C}$ classifier, and let $x_{i}\in X$ be a well-classified image by $\phi$. Then, assuming that $\phi_{F}(x_{i}) = \overrightarrow{f}^{i}$,
\begin{itemize}
    \item $N_{\phi_{C}} = \{n \ | \ n \in\phi_{C} \ a \ neuron\} $ is the set of neurons of $\phi_{C}$.
    \item $E^{x_{i}}_{\phi_{C}} = \{nn'\ | \ n,n'\in N_{\phi_{C}} \ and \ nn' \ is \ the \ connection \ $ \\ $between \ n \ and \ n'\}$ is the set of connections of $\phi_{C}$ in $x_{i}$.
    \item $A^{x_{i}}_{\phi_{C}} = \{a_{nn'}\ | \ n,n' \in N_{\phi_{C}} \ and \ a_{nn'} \ is \ the \ activation \ $ \\ $ associated \ with \ nn'\in E^{x_{i}}_{\phi_{C}} \ when \ \phi_{C}(\overrightarrow{f}^{i})\}$ is the set of activations of $\phi_{C}$ in $x_{i}$.  
\end{itemize}

Those sets define a Weighted Digraphs \cite{wilson1979introduction}, $G^{x_{i}}_{\phi_{C}} = (N_{\phi_{C}}, E^{x_{i}}_{\phi_{C}}, A^{x_{i}}_{\phi_{C}})$, where the nodes of the graph are the neurons of $\phi_{C}$ and the weighted edges of the graph are given by the activations of the $\phi_{C}$. $G^{x_{i}}_{\phi_{C}}$ is the behavior graph of $\phi_{C}$ in $x_{i}$.    

\end{definition}

As far as we know, the first example in this field of this type of graph was presented by Echeberria et al. in \cite{echeberria2022understanding}, which is used to visualize the behavior of the targeted model processing an input sample. 

\subsection{Classifier Node Attributes}\label{subsec:attributes}

Now, through a classifier's behavior graph, several attributes of the $\phi_{C}$'s neurons can be computed. In this section, those node characteristics are presented and detailed. Notice that each image $x_{i}\in X$ gives a classifier's behavior graph $G^{x_{i}}_{\phi_{C}}$, which will be associated with a set of neuron attributes $\Delta^{X}_{\phi_{C}}$. Those neuron attributes are called: \textit{Impact}, \textit{Influence}, \textit{Input Proportion}, and \textit{Specialization}.    

\subsubsection{Impact}

This neuron attribute is recovered from the scientific article by Echeberria et al. \cite{echeberria2022understanding}, where it was defined as the difference between the output of the targeted neuron and the addition of its inputs. However, in this study, this definition is modified slightly.

Observe that the classifier can contain two layers with different activation functions. Therefore, the impact attribute compares two numbers on different scales since the output and the input activations could not share the scale. A clear example is when a layer contains the relu function, and the next layer includes the sigmoid function. In the comparison, the sigmoid activations can finally be insignificant because the sigmoid function is bounded by one, while the relu is unbounded. Echeberria et al. \cite{echeberria2022understanding} do not consider that setback due to their targeted model only contains a unique activation function, relu. Hence, in this work, the slight modification of the impact feature incorporates an operation that normalizes the activations for each classifier's layer.

Suppose $a^{l}_{1}, a^{l}_{2}, ...,a^{l}_{s_{l}}\in A^{x_{i}}_{\phi_{C}}$ are the activation values of the layer $\phi^{l}_{C}$, when $\phi$ computes $x_{i}$. Note that $s_{l}$ is the number of neurons in the layer $\phi^{l}_{C}$. Scaling those values by layer,
\begin{equation}\label{eq:normalizaion}
\alpha^{l}_{j} = \frac{a^{l}_{j}}{\sum^{s_{l}}_{k = 1}a^{l}_{k}}
\end{equation}
, where $j \in \{1, ..., s_{l}\}$. Concretely, $0\leq\alpha^{l}_{j}\leq 1$ for whatever layer's activation. Therefore, assuming that $\mathcal{A}^{x_{i}}_{\phi_{C}}$ is the set of those normalized activation values by layer, working with $\mathcal{A}^{x_{i}}_{\phi_{C}}$ the problem with the scale are solved.

\begin{definition}[Impact]

Let $\phi$ be a deep learning model containing $\phi_{F}$ feature-extractor block and $\phi_{C}$ classifier, and let $x_{i}\in X$ be a well-classified image by $\phi$. Therefore, if $G^{x_{i}}_{\phi_{C}} = (N_{\phi_{C}}, E^{x_{i}}_{\phi_{C}}, A^{x_{i}}_{\phi_{C}})$ is the behavior graph of $\phi_{C}$ in $x_{i}$,
\begin{equation*}
    \mathcal{A}^{x_{i}}_{\phi_{C}} = \{\alpha_{nn'} \ | \ \alpha_{nn'}= \frac{a_{nn'}}{\sum\limits_{\substack{k\in \phi^{l}_{C}}} a_{kn'}},\ \text{where} \ n\in \phi^{l}_{C} \ \text{and} \ a_{nn'},a_{kn'}\in A^{x_{i}}_{\phi_{C}}\}
\end{equation*}

is the set of normalized activations of $\phi_{C}$ in $x_{i}$. Moreover, let $n$ a neuron of the layer $\phi^{l}_{C}$ and $n'$ any neuron of the $\phi^{l+1}_{C}$, then 

$$ \iota^{x_{i}}_{n} = \alpha_{nn'} - \frac{\sum\limits_{\substack{N\in \phi^{l-1}_{C}}} \alpha_{Nn}}{|\phi^{l-1}_{C}|} $$

is the impact of the $\phi_{C}$'s neuron $n$ in $x_{i}$. Notice that, does not matter the $n'$ neuron of the definition, since $\alpha_{nn'}$ is the same activation for all $n'$. 

\end{definition}

This feature shows the impact of a neuron in the prediction of an image in the sense of how it is modifying the values that go through the classifier. Imagine the neuron $n\in \phi_{C}$ with the impact $\iota^{x_{i}}_{n}$ in $x_{i}$. Table \ref{tab:interpretations} shows the five main interpretations of this attribute according to the value it takes.




\begin{table}[!ht]
\centering
\caption{The five main interpretations of the impact feature according to the value range.}
\label{tab:interpretations}
\resizebox{0.48\textwidth}{!}{
\begin{tabular}{l|l}
\hline\noalign{\smallskip}
Value range & Interpretation\\
\noalign{\smallskip}\hline\noalign{\smallskip}
$\iota^{x_{i}}_{n} >> 0$ & $n$ modifies highly and positively the received values on the prediction of the $x_{i}$\\
\noalign{\smallskip}\hline\noalign{\smallskip}
$\iota^{x_{i}}_{n} > 0$ & $n$ modifies slightly and positively the received values on the prediction of the $x_{i}$\\
\noalign{\smallskip}\hline\noalign{\smallskip}
$\iota^{x_{i}}_{n} = 0$ & $n$ kepdf the received values equal on the prediction of the $x_{i}$ \\
\noalign{\smallskip}\hline\noalign{\smallskip}
$\iota^{x_{i}}_{n} < 0$ & $n$ modifies slightly and negatively the received values on the prediction of the $x_{i}$.\\
\noalign{\smallskip}\hline\noalign{\smallskip}
$\iota^{x_{i}}_{n} << 0$ & $n$ modifies highly and negatively the received values on the prediction of the $x_{i}$\\
\noalign{\smallskip}\hline
\end{tabular}}
\end{table}

\begin{figure}[!ht]
    \centering
    \includegraphics[width=0.46\textwidth]{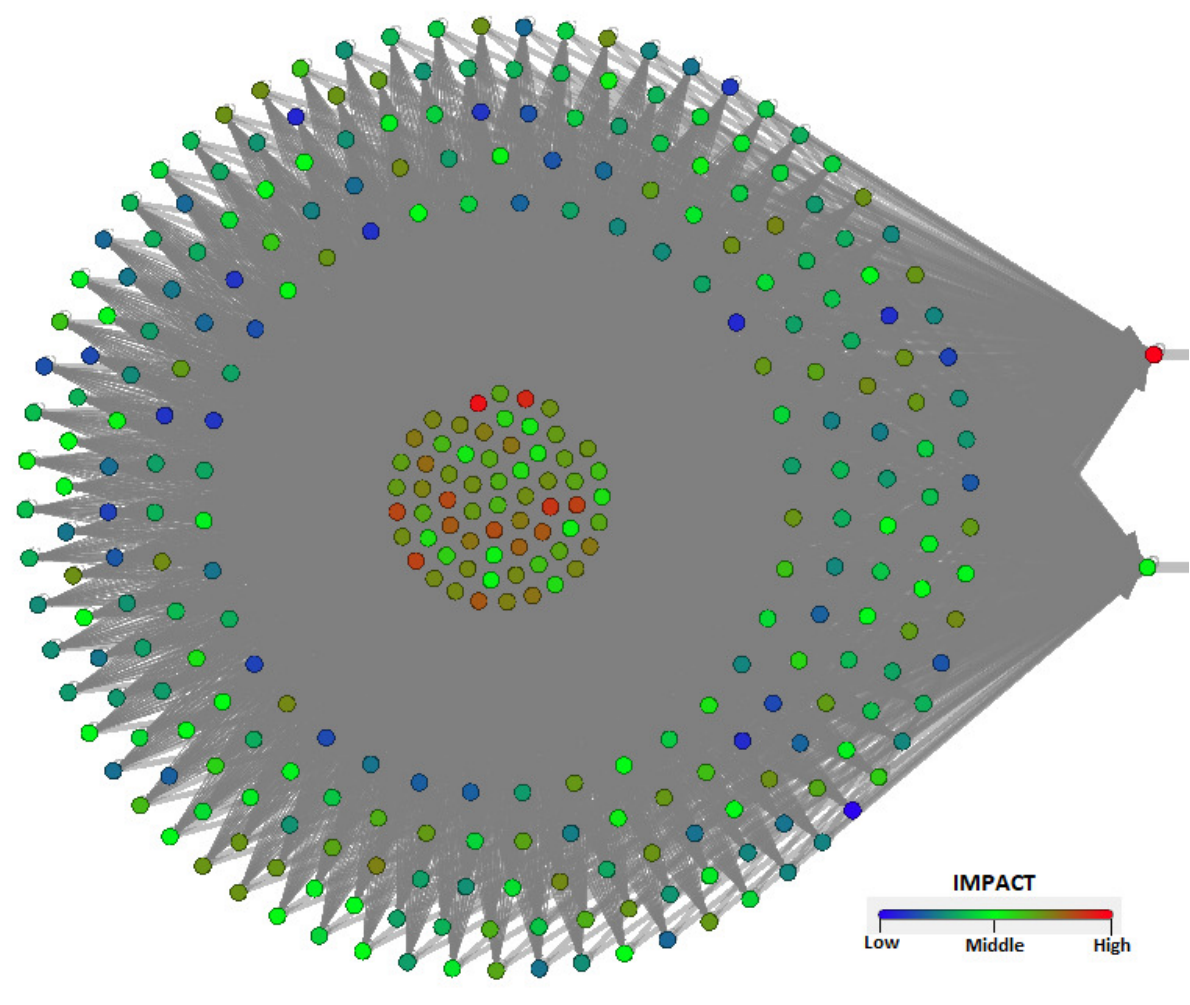}
    \caption{Classifier Behavior Graph example, where the nodes are coloring according to their impact attribute}
    \label{fig:impact_vis}
\end{figure}

It is possible to visualize a classifier behavior graph $G^{x_{i}}_{\phi_{C}}$ coloring the nodes according to the associated impact attribute. Concretely, Figure \ref{fig:impact_vis} shows the impact of each neuron, where blue neurons impact negatively on the prediction while the red neurons impact positively. The green neurons' impact is nearly null.

\subsubsection{Influence}

This node attribute was presented by Hohman et al. \cite{8807294}, where it is used to discover essential filters of the convolutional neural network in predicting images. The presented definition is modified in this work to implement it in a dense neural network, that is, in the classifier of the targeted model. Therefore, the presented new influence attribute highlights the neurons per layer that participate in the prediction with the highest activation values. The activation value can be understood as how influential the associated neuron is in the classification; hence, this attribute is named influence.   

This attribute considers the target neuron $n\in \phi_{C}$, the neuron's activation value $a_{nn'}\in A^{x_{i}}_{\phi{C}}$ in the image $x_{i}\in X$, and the layer where the neuron is, $\phi^{l}_{C}$. Suppose that $\{a^{l}_{1}, ..., a^{l}_{s_{l}}\}\subset A^{x_{i}}_{\phi{C}}$ are the activations of the neurons in $\phi^{l}_{C}$. Those values may lose control in the sense of the scale; for example, in case of the layer's activation is relu, the scale of the obtained values can increase considerably. Therefore, the activations are normalized by the equation presented in (\ref{eq:normalizaion}).   

Once the preprocess is finished, the set $\{\alpha^{l}_{1}, ..., \alpha^{l}_{s_{l}}\}$ of the normalized activations of the neurons in $\phi^{l}_{C}$ is obtained, where $0\leq \alpha^{l}_{j} \leq 1$ for all $ j \in \{1, ..., s_{l}\}$ and $\sum^{s_{l}}_{j = 1} \alpha^{l}_{j} = 1$. Moreover, each $\alpha^{l}_{j}$ is associated with a neuron in the layer number $l$ of the classifier $\phi_{C}$, that is, concretely with $n^{l}_{j} \in \phi^{l}_{C}$. Now, the set of the normalized activations is sorted, obtainining $\{\alpha^{l}_{\sigma(1)}, ..., \alpha^{l}_{\sigma(s_{l})}\}$, where $\sigma$ is the permutation that sort the set from the highest to the lowest. Assuming the $0\leq p\leq 1$ parameter, the normalized activations are added till obtain a value equal or higher than $p$ 
\begin{equation}
\sum_{j = 1}^{\sum_{j} \alpha^{l}_{\sigma(j)}\geq p} \alpha^{l}_{\sigma(j)} = p' \leq p
\end{equation}   
Suppose that $\alpha^{l}_{\sigma(1)}, ..., \alpha^{l}_{\sigma(s'_{l})}$ are the values participating in the summation. Therefore, the influence attribute of the $n^{l}_{\sigma(1)}, ..., n^{l}_{\sigma(s'_{l})}$ is $1$ while the influence of the others is $0$.

\begin{definition}[Influence]

Let $\phi$ be a deep learning model containing $\phi_{F}$ feature-extractor block and $\phi_{C}$ classifier, and let $x_{i}\in X$ be a well-classified image by $\phi$. Moreover, let $G^{x_{i}}_{\phi_{C}} = (N_{\phi_{C}}, E^{x_{i}}_{\phi_{C}}, A^{x_{i}}_{\phi_{C}})$ be the behavior graph of $\phi_{C}$ in $x_{i}$, and let 
\begin{equation*}
    \mathcal{A}^{x_{i}}_{\phi_{C}} = \{\alpha_{nn'} \ | \ \alpha_{nn'}= \frac{a_{nn'}}{\sum\limits_{\substack{k\in \phi^{l}_{C}}} a_{kn'}},\ \text{where} \ n\in \phi^{l}_{C} \ \text{and} \ a_{nn'},a_{kn'}\in A^{x_{i}}_{\phi_{C}}\}
\end{equation*}
be the set of normalized activations of $\phi_{C}$ in $x_{i}$. Therefore, if $n \in \{n_{1}, ..., n_{s_{l}}\} = \phi^{l}_{C}$, $n'$ is any neuron of the $\phi^{l+1}_{C}$, $\sigma$ is the permutation that sort $\{\alpha_{n_{\sigma(1)}n'}, ..., \alpha_{n_{\sigma(s_{l})}n'}\}\subset \mathcal{A}^{x_{i}}_{\phi{C}}$ from greatest to lowest, and $0\leq p\leq 1$ is a parameter, then 
\begin{equation*}
    \xi^{x_{i}, p}_{n} =
    \begin{cases}
      1, & \text{if}\ n \in \{n_{\sigma(1)}, ..., n_{\sigma(j_{p})}\}, \ \text{where}\ \sum\limits_{j = 1}^{j_{p}} \alpha_{n_{\sigma(j)}n'} \leq p < \sum\limits_{j = 1}^{j_{p}+1} \alpha_{n_{\sigma(j)}n'}\\
      0, & \text{if}\ n \in \{n_{\sigma(j_{p}+1)}, ..., n_{\sigma(s_{l})}\}, \ \text{where}\ \sum\limits_{j = 1}^{j_{p}} \alpha_{n_{\sigma(j)}n'} \leq p < \sum\limits_{j = 1}^{j_{p}+1} \alpha_{n_{\sigma(j)}n'} \\
    \end{cases}
\end{equation*}
is the influence of the $\phi_{C}$'s neuron $n$ in $x_{i}$. Notice that, does not matter the $n'$ neuron of the definition, since $\alpha_{nn'}$ is the same activation for all $n'$. 
\end{definition}

\begin{figure}[!ht]
    \centering
    \includegraphics[width=0.46\textwidth]{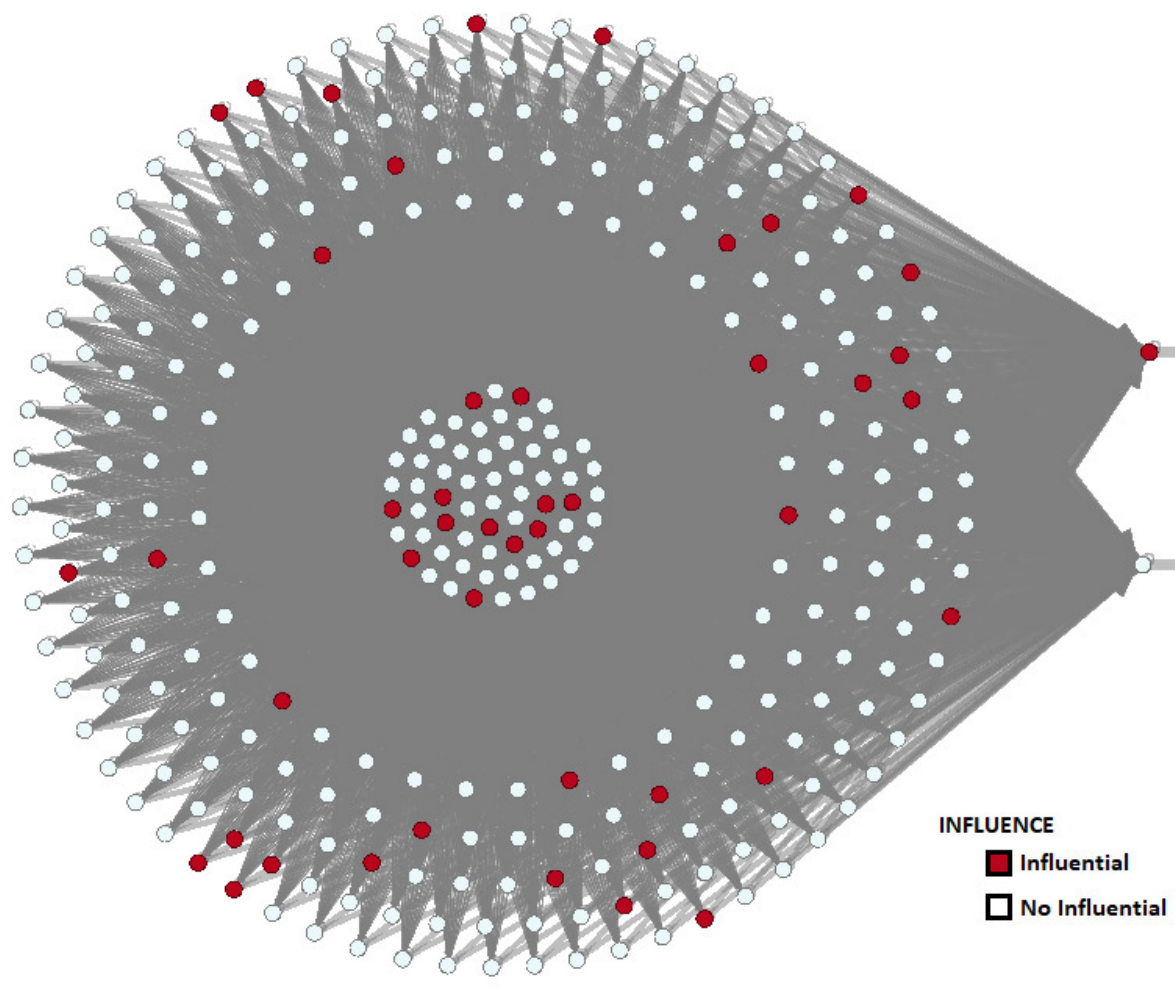}
    \caption{Classifier Behavior Graph example, where the nodes are coloring according to their influence attribute}
    \label{fig:influence_vis}
\end{figure}

This attribute highlights the most influential neurons in a layer. In other words, this characteristic takes $1$ as a value when the neuron's participation is influential enough according to the $p$ parameter and respects the rest neurons of the same layer. It is possible to visualize a classifier behavior graph $G^{x_{i}}_{\phi_{C}}$ coloring the highlighted nodes according to the associated influence attribute. Concretely, Figure \ref{fig:influence_vis} shows the influence of each neuron, where the highlighted neurons are colored in red and the others are not.

\subsubsection{Input Proportion}

This attribute is not obtained from another scientific article, so it is proposed in this work as a possible attribute. It computes the proportion of the input value number that a neuron receives with respect to all the input values it can receive according to the topology of the targeted model. In other words, it is the percentage of the no null input values received.  

This attribute is implemented in the classifier of the targeted model. Note that this attribute gives information about the targeted layer according to the activations of the previous layer. Consider $\phi$ a model classifying cats and dogs images, $n\in\phi^{l}_{C}$ a neuron in layer $l$, $\{n_{1},...,n_{s_{l-1}}\}=\phi^{l-1}_{C}$ all the neurons in the layer $l-1$, and $\{a^{x_{i}}_{n_{1}n},..., a^{x_{i}}_{n_{s_{l-1}}n}\}\subseteq A^{x_{i}}_{\phi_{C}}$ are the activation values that are received by $n$ when $\phi$ computes $x_{i}$. Imagine  $x_{dog}$ and $x'_{dog}$ are dog images and $x_{cat}$ is cat images, then it is reasonable think that 
$$\{a^{x_{dog}}_{n_{1}n},..., a^{x_{dog}}_{n_{s_{l-1}}n}\}\approx \{a^{x'_{dog}}_{n_{1}n},..., a^{x'_{dog}}_{n_{s_{l-1}}n}\}$$
, while
$$ \{a^{x_{dog}}_{n_{1}n},..., a^{x_{dog}}_{n_{s_{l-1}}n}\}\not\approx\{a^{x_{cat}}_{n_{1}n},..., a^{x_{cat}}_{n_{s_{l-1}}n}\}$$   
Particularly, the null activation values would be similar. Mathematically, 
$$\sum\limits_{a^{x_{dog}}_{n_{j}n}\neq 0} 1 \approx \sum\limits_{a^{x'_{dog}}_{n_{j}n}\neq 0} 1 \not\approx \sum\limits_{a^{x_{cat}}_{n_{j}n}\neq 0} 1$$ 
For example, if an activation function is related to if in the image there is a snout or not. It will be null in all the cat images due to every dog has a snout while cats do not.

Suppose an adversary wants to classify a cat image as a dog modifying the image slightly. Assume $x'_{cat} = x_{cat}+\epsilon$ a corrupted cat image that the target model classifies as a dog. The little modification generates a few perturbations in the activation values generated by $x_{cat}$. That is,
$$\{a^{x_{cat}}_{n_{1}n},..., a^{x_{cat}}_{n_{s_{l-1}}n}\}\approx \{a^{x'_{cat}}_{n_{1}n},..., a^{x'_{cat}}_{n_{s_{l-1}}n}\}\Rightarrow \sum\limits_{a^{x_{cat}}_{n_{j}n}\neq 0} 1\approx \sum\limits_{a^{x'_{cat}}_{n_{j}n}\neq 0} 1\not\approx \sum\limits_{a^{x_{dog}}_{n_{j}n}\neq 0} 1$$
but $x'_{cat}$ is classified as dog. Therefore, the targeted model classifies an image as a dog that generates a null activation similar to the average cat image (weird for an dog image).

\begin{definition}[Input Proportion]

Let $\phi$ be a deep learning model containing $\phi_{F}$ feature-extractor block and $\phi_{C}$ classifier, and let $x_{i}\in X$ be a well-classified image by $\phi$. Moreover, let $G^{x_{i}}_{\phi_{C}} = (N_{\phi_{C}}, E^{x_{i}}_{\phi_{C}}, A^{x_{i}}_{\phi_{C}})$ be the behavior graph of $\phi_{C}$ in $x_{i}$. Therefore, if $n \in \phi^{l}_{C}$ and $n_{1}, ..., n_{s_{l-1}} \in \phi^{l-1}_{C}$ are all the neurons that form $\phi^{l-1}_{C}$, then 
\begin{equation*}
    \rho^{x_{i}}_{n} = \frac{\sum\limits_{a_{n_{j}n}\neq 0} 1}{\sum\limits_{a_{n_{j}n}} 1}
\end{equation*}
where $a_{n_{j}n}\in\{a_{n_{1}n}, ..., a_{n_{s_{l}}n}\}\subset A^{x_{i}}_{\phi_{C}}$, is the input proportion of the $\phi_{C}$'s neuron $n$ in $x_{i}$.   
\end{definition}

This attribute depends on the targeted model's activation function, giving more or less information according to it. The sigmoid activation function never returns a null value due to its definition, while the relu activation function maps several values to zero. This attribute takes into account the null values; therefore, in the case of the activation function that never annuls values, it does not give any information since all the images generate the same number of null activations, i.e., no null activation. However, in the case of the activation function maps values to zero, yes, it gives information.    

\begin{figure}[!ht]
    \centering
    \includegraphics[width=0.46\textwidth]{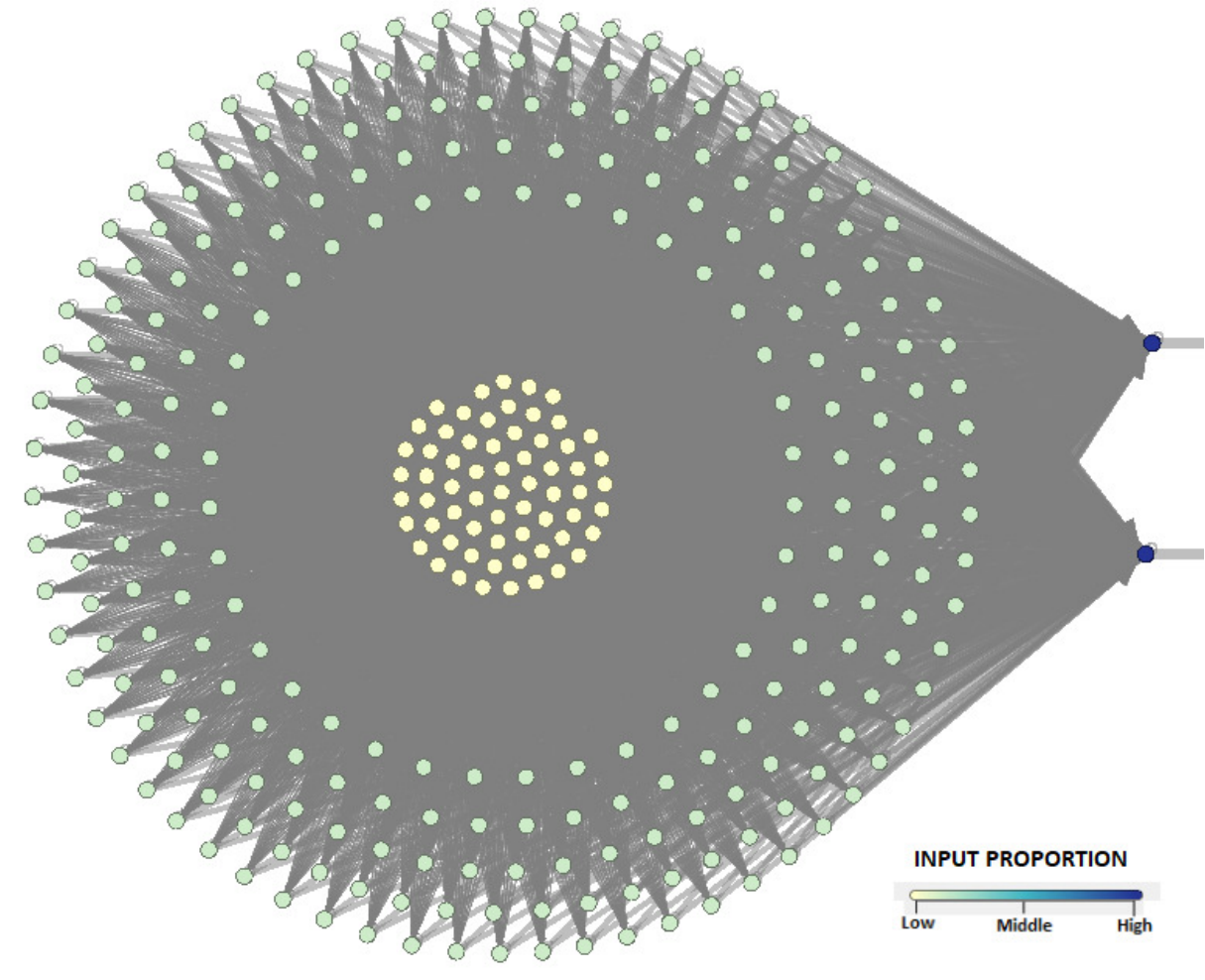}
    \caption{Classifier Behavior Graph example, where the nodes are coloring according to their input proportion attribute}
    \label{fig:input_prop_vis}
\end{figure}

Continuing with the visualization of the generated attributes, Figure \ref{fig:input_prop_vis} shows a classifier behavior graph $G^{x_{i}}_{\phi_{C}}$ coloring the nodes according to the input proportion attribute. Note that all neurons of a concrete layer have associated with the same input proportion value since the classifier part of the targeted model is a fully connected model. In other words, an activation value of a certain neuron in the classifier is passed to all the neurons of the next classifier's layer. That is why Figure \ref{fig:input_prop_vis} shows three different colors, one of them for each layer that forms the targeted model's classifier. Observe that the input layer neurons have the minimum input proportion, followed by the neurons of the hidden layer that contains a bigger value associated with this attribute. Finally, the maximum input proportion value is given to the two neurons that form the output layer. 

\subsubsection{Specialization}

The idea of this node attribute is obtained from the paper presented by Echeberria et al. \cite{echeberria2022understanding}, where a similar term is introduced to visualize and understand the modification of the behavior graph in the process of obtaining an adversarial example of an image. Concretely, they check the participation of the neurons predicting the different classes, computing the frequency of each neuron. Then, among all the neurons, they select the most polarized ones, i.e., the neurons that participate much more or considerably more in predicting a concrete class. Note that the desired neurons are the neurons with a high frequency in a unique class prediction. Finally, they show that those neurons are particularly modified, as is expected, in the obtention of an adversarial example. Therefore, this work considers this presented term as a node attribute, which computes the mentioned frequency, but its definition is modified slightly.
     
The modification is related to the meaning that a neuron does not participate in the prediction. Echeberria et al. \cite{echeberria2022understanding} consider that a neuron does not participate in a prediction when its activation value is zero. This definition makes sense when the target model is formed by activation functions such as relu that maps some values to zero. However, in the case of other activation functions, such as sigmoid, that definition does not have sense because there will not be a null activation value. Therefore, this attribute considers that a neuron does not participate in the prediction when its activation value is not among the $k$-top activation values. Moreover, the highlighted neurons are selected per layer, i.e., the desired neurons are the $k$-top from each layer.

That attribute is divided into many sub-attributes given by the number of classes in the targeted dataset. Suppose the targeted dataset contains two classes, $0$ and $1$. In that case, there will be two specialization attributes, specialization $0$ and specialization $1$, representing the frequency in the participation predicting the class $0$ and class $1$, respectively. 

First, for this attribute is necessary to compute several sets, which will be called the set of the specialization values in class $\mathfrak{c}$. Then, each class $\mathfrak{c}$ gives a set containing values associated with the neurons of the classifier, which will indicate the neurons' specialization predicting the class $\mathfrak{c}$. Therefore, suppose that $\mathfrak{c}$ a class, $X_{\mathfrak{c}}\subseteq X$ the images of that class, and $k$ a parameter, then
\begin{equation*}
    \vartheta^{x_{i}, k}_{nn'} =
    \begin{cases}
      1, & \text{if}\ a_{nn'}\in A^{x_{i}}_{\phi_{C}}\ \text{is top}\ k \ \text{in} \ \{a_{n_{1}n'}, ..., a_{n_{s_{l}}n'}\}\subset A^{x_{i}}_{\phi_{C}}, \ \text{where}\ n \in \{n_{1}, ..., n_{s_{l}}\} = \phi^{l}_{C}\\
      0, & \text{else}\\
    \end{cases}
\end{equation*}
and
$$\mathcal{S}^{\mathfrak{c}, k}_{\phi_{C}} =  \left\{\zeta^{\mathfrak{c},k}_{nn'}\ | \ \zeta^{\mathfrak{c},k}_{nn'}= \frac{\sum\limits_{x_{i}\in X_{\mathfrak{c}}}\vartheta^{x_{i}, k}_{nn'}}{|X_{\mathfrak{c}}|}\right\}$$
is the set of the classifier behavior graphs of $\phi_{C}$ in the images of class $\mathfrak{c}$. Notice that those values indicate how many times the activation of the node $n$ is top $k$ predicting the images in class $\mathfrak{c}$. Even though a neuron $n$ can contain more than one output connection if in the next layer there are more that one neuron, they have the same value associated. In other words, if there exist $n'$ and $n''$ two different neurons in $\phi_{C}^{l+1}$ that $\zeta_{nn'}, \zeta_{nn''} \in \mathcal{S}^{\mathfrak{c}, k}_{\phi_{C}}$, then $\zeta_{nn'} = \zeta_{nn''}$ because $\forall n \in \phi_{C}^{l}\ a_{nn'} = a_{nn''}$.       

Once the set of the classifier behavior graphs of $\phi_{C}$ in the images of class $\mathfrak{c}$ is presented, it is possible to define the specialization $\mathfrak{c}$ attribute. 

\begin{definition}[Specialization $\mathfrak{c}$]

Let $\phi$ be a deep learning model containing $\phi_{F}$ feature-extractor block and $\phi_{C}$ classifier, and let $x_{i}\in X$ be a well-classified image by $\phi$. Moreover, let $G^{x_{i}}_{\phi_{C}} = (N_{\phi_{C}}, E^{x_{i}}_{\phi_{C}}, A^{x_{i}}_{\phi_{C}})$ be the behavior graph of $\phi_{C}$ in $x_{i}$, let $k$ be a parameter, and let $\mathcal{S}^{\mathfrak{c}, k}_{\phi_{C}}$ be the set of the classifier behavior graphs of $\phi_{C}$ in the images of class $\mathfrak{c}$. Therefore, if $n\in \phi^{l}_{C}$ and $n'$ is any neuron in $\phi^{l+1}_{C}$, then 
\begin{equation*}
    \zeta^{x_{i}, \mathfrak{c}, k}_{n} = \begin{cases}
      \zeta^{\mathfrak{c},k}_{nn'}, & \text{if}\ nn'\in E^{x_{i}}_{\phi_{C}}\ \text{and}\ a_{nn'}\neq 0\\ 
      0, & \text{if}\ nn'\in E^{x_{i}}_{\phi_{C}}\ \text{and}\ a_{nn'} = 0\\
    \end{cases}
\end{equation*}
where $a_{nn'}\in A^{x_{i}}_{\phi_{C}}$, is the specialization $\mathfrak{c}$ of the $\phi_{C}$'s neuron $n$ in $x_{i}$.   
\end{definition}

\begin{figure}[!ht]
  \centering
  \begin{subfigure}[b]{0.23\textwidth}
    \includegraphics[width=\textwidth]{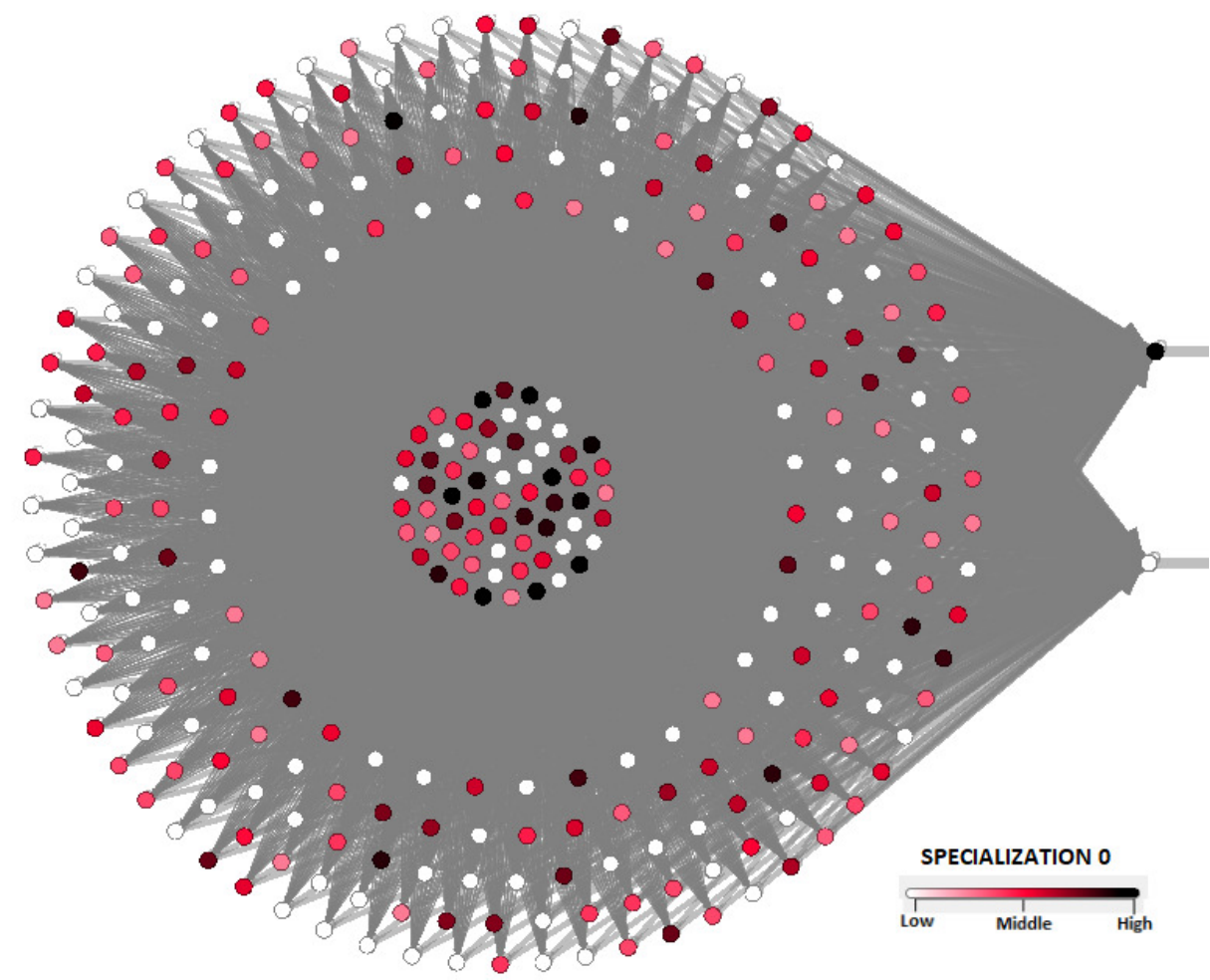}
    \caption{Specialization $0$}
    \label{fig:specialization_0_vis}
  \end{subfigure}
  \begin{subfigure}[b]{0.23\textwidth}
    \includegraphics[width=\textwidth]{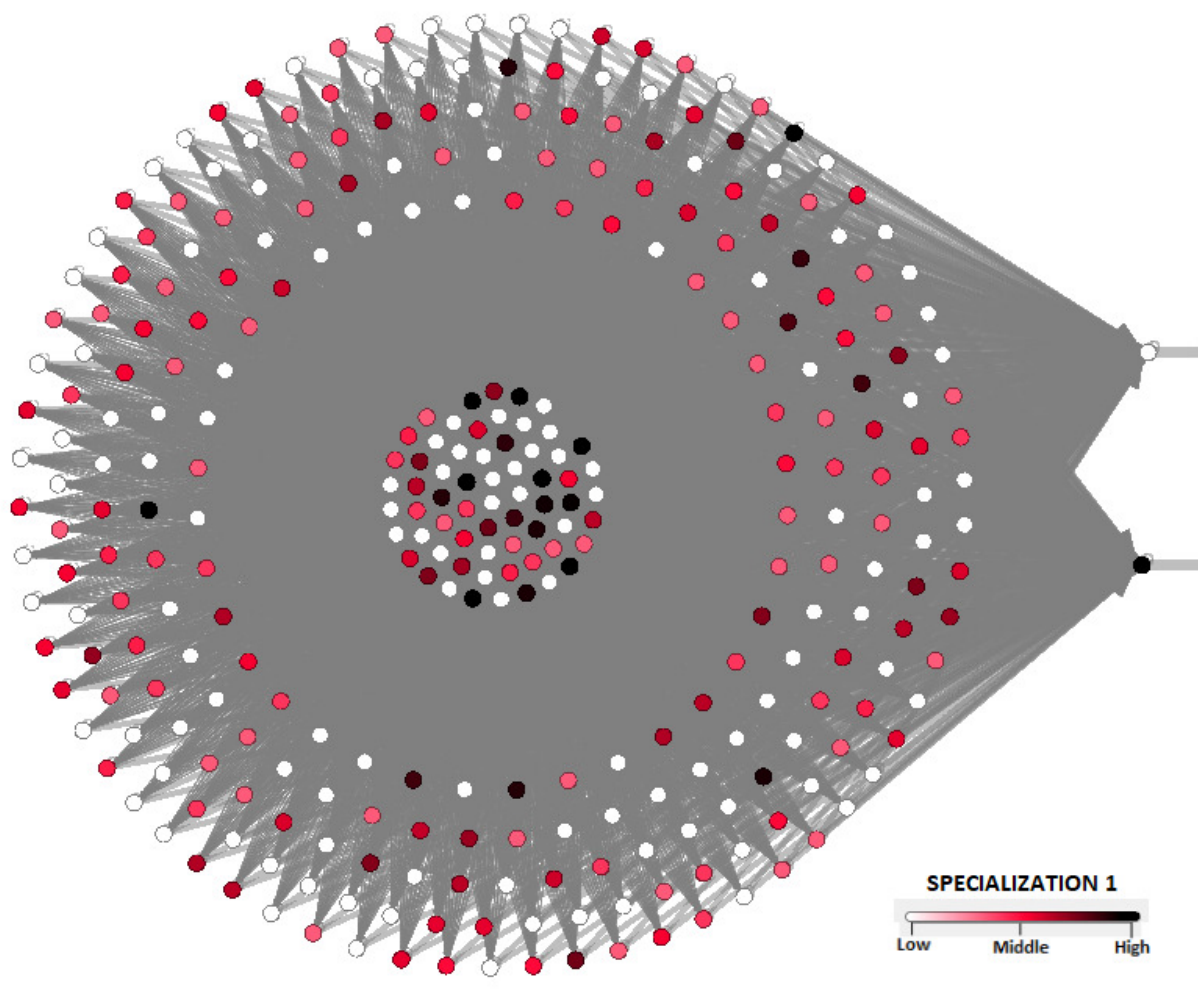}
    \caption{Specialization $1$}
    \label{fig:specialization_1_vis}
  \end{subfigure}
  \caption{Classifier Behavior Graph example, where the nodes are coloring according to their specialization attributes}
  \label{fig:specialization_vis}
\end{figure}

This attribute gives information about the activity, predicting class $\mathfrak{c}$ images, of the neurons participating in the classification of $x_{i}$. Therefore, a node attribute is obtained for each class $c$ from the dataset. Figure \ref{fig:specialization_vis} shows the visualization of the same classifier behavior graph but with two different colorizations. In this case, the dataset would be composed of two classes, and each graph shows the specialization of one of them. In both cases, the less specialized neurons are colored with white color, while the highest specialized are the black ones. Finally, the neurons associated with a middle-value of specialization are colored in red. This combination of colors generates a scale of white-red-black, indicating the specialization of a neuron.     

\subsection{Data preprocessing}
\label{subsec:preprocess}

This subsection presents data preprocessing, which gives an enriched dataset from a dataset and the targeted model. It is formed by three subsets defined below. Remark that the preprocessed dataset is used to develop the evasion attack detector.

Suppose that $\mathcal{X}$ is a set of adversarial and original images of a dataset, $\mathcal{X}$ is formed by $\{\mathfrak{c}_{0},...,\mathfrak{c}_{t-1}\}$ image classes, and $\phi$ is the targeted model composed of a feature-extractor block, $\phi_{F}$, and the classifier, $\phi_{C}$. Now, the preprocess of an image $x\in \mathcal{X}$ is the following:
\begin{enumerate}
	\item $G^{x}_{\phi_{C}} = (N_{\phi_{C}}, E^{x}_{\phi_{C}}, A^{x}_{\phi_{C}})$, the behavior graph of $\phi_{C}$ in $x$ is computed.
	\item $Ad^{x}_{\phi_{C}}$, the adjacency matrix of $G^{x}_{\phi_{C}}$ is obtained, which is defined as
\begin{equation*}
a_{n_{i}n{j}} = \begin{cases}
      1, & \text{if}\ n_{i}n_{j} \in E^{x_{i}}_{\phi_{C}} \\
      0, & \text{if}\ n_{i}n_{j}\not\in E^{x_{i}}_{\phi_{C}}. \\
    \end{cases} 
\end{equation*}
	\item $I^{x}_{ \phi_{C}} = \{\iota^{x}_{n}\ | \ \forall n \in \phi_{C}\}$, the set of the impact attributes of the neurons of $\phi_{C}$ in $x$ is calculated.
	\item $\Xi^{x}_{ \phi_{C}} = \{\xi^{x,p}_{n}\ | \ \forall n \in \phi_{C}\}$, the set of the influence attributes of the neurons of $\phi_{C}$ in $x$ is computed.
	\item $P^{x}_{ \phi_{C}} = \{\rho^{x}_{n}\ | \ \forall n \in \phi_{C}\}$, the set of the input proportion attributes of the neurons of $\phi_{C}$ in $x$ is obtained.
	\item $Z^{x, \mathfrak{c}_{i}, k}_{ \phi_{C}} = \{\zeta^{x, \mathfrak{c}_{i}, k}_{n}\ | \ \forall n \in \phi_{C}\}$, the set of the specialization $\mathfrak{c}_{i}$ attributes of the neurons of $\phi_{C}$ in $x$ is calculated, for all $i\in\{0,1,...,t-1\}$.
	\item Finally, grouping and reorganizing those sets of attributes, the set of neuron attributes of $\phi_{C}$ in $x$ is obtained: 
    \begin{equation}
	\Delta^{x, p, k, t}_{\phi_{C}} = \{\delta^{x, p, k, t}_{n} = (\iota^{x}_{n}, \xi^{x,p}_{n}, \rho^{x}_{n}, \zeta^{x, \mathfrak{c}_{i}, k}_{n},..., \zeta^{x, \mathfrak{c}_{t-1}, k}_{n})\ | \ \forall n \in \phi_{C}\} 
    \end{equation}
\end{enumerate}

Therefore, for each image $x\in \mathcal{X}$, both $\Delta^{x, p, k}_{\phi_{C}}$ and $Ad^{x}_{\phi_{C}}$ are obtained. Moreover, a new label $\lambda^{x}_{\phi_{C}}$ is generated, where 
\begin{equation*}
\lambda^{x}_{\phi_{C}} = \begin{cases}
      1, & \text{if\ is \ an \ adversarial\ example} \\
      0, & \text{if\ is\ not \ an \ adversarial\ example}. \\
    \end{cases} 
\end{equation*}
is the label of $x$ in $\phi_{C}$ for the new evasion attack detector.

This preprocess is repeated for each image $x$ in $\mathcal{X}$, expanding all the defined sets to whole $\mathcal{X}$. Hence, $(\Delta(\mathcal{X})^{p, k, t}_{\phi_{C}}, Ad(\mathcal{X})_{\phi_{C}})$ and $\Lambda(\mathcal{X})_{\phi_{C}}$ is the new dataset to work, where
\begin{align}
\label{eq:set1}
    &\Delta(\mathcal{X})^{p, k, t}_{\phi_{C}} = \{\Delta^{x, p, k, t}_{\phi_{C}}\ |\ x\in \mathcal{X}\}\\
\label{eq:set2}
    &Ad(\mathcal{X})_{\phi_{C}} = \{Ad^{x}_{\phi_{C}} \ |\ x\in \mathcal{X}\}\\
\label{eq:set3}
    &\Lambda(\mathcal{X})_{\phi_{C}} = \{\lambda^{x}_{\phi_{C}} \ |\ x\in \mathcal{X}\}
\end{align}

\subsection{Evasion attack detector}
\label{subsec:detector}

The section of the detector's technology is based on the idea that the topology of the deep learning models, particularly the classifier's topology, is essential to predict the data. In other words, the topological position of the anomalous neuron activity in the classifier is essential information to maintain. Therefore, graph neural networks are the selected technology to develop the detector. Concretely, the detector is based on the graph convolutional network (GCN) technology \cite{kipf2017semisupervised}, which gives outstanding results compared with other types of graph neural network technology \cite{zhang2018graph, 9601152}. 

\begin{figure}[!ht]
    \centering
    \includegraphics[width=0.46\textwidth]{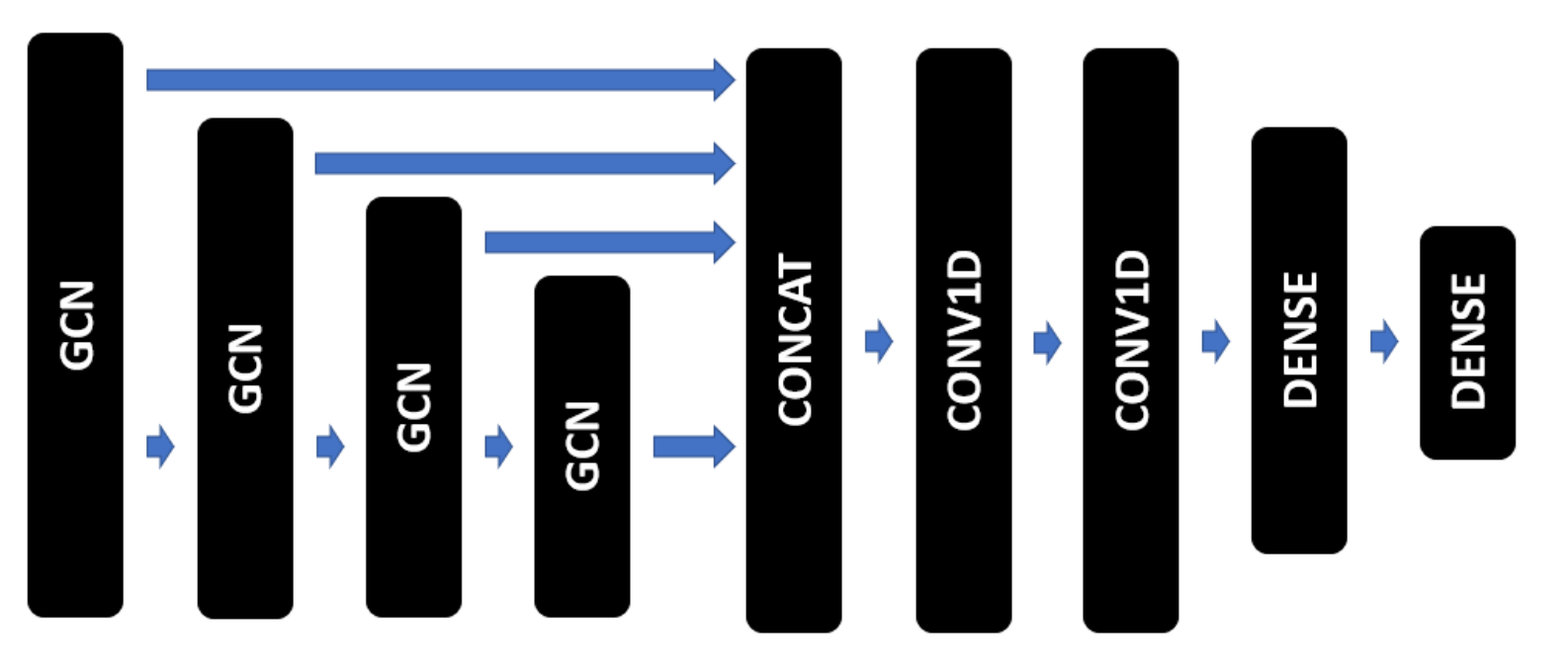}
    \caption{The selected architecture for developing the evasion attack detector}
    \label{fig:detector_architecture}
\end{figure}

Otherwise, several deep learning GCN models with different architectures can be found in the literature \cite{zhang2019graph}. However, the desired architecture is obtained from \cite{DBLP:journals/corr/abs-1902-06673}, which obtains noted results detecting fake news. Therefore, the selected detector's architecture is shown in Figure \ref{fig:detector_architecture}. Even though this architecture is proposed, others may be considered in the future. Due to the novelty of the proposed detector, there was not enough study on this aspect to obtain the optimal architecture. Therefore, future works may be analyzed deeply different architectures, improving this novel type of evasion attack detector.  

The detector begins with a block of GCN layers, where all of them contain the hyperbolic tangent activation function. Moreover, the first three GCN layers are formed by $32$ filters, while the last one is composed of a unique filter. Notice that all the GCN layer outputs are concatenated and transferred to another block containing one-dimensional convolution layers \cite{perez2019guide}. Notice that any participating neuron has $97$ associated attributes at the convolutional block since it received $32$ attributes in the first three GCN layers and one in the last. Because of that, this architecture implements the first convolution layer with $16$ filters of size $97$, taking stpdf of length $97$ to study each node by separating. The next convolution layer tries to obtain information at the graph level, and it is implemented with $32$ filters of size $5$. Both convolutional layers contain the relu activation function. Finally, the classifier receives the information from the convolution block. It is formed by two dense layers containing the sigmoid activation function. The first dense layer is formed by $200$ neurons, while the output layer comprises two neurons. 

\section{Experiment}
\label{sec:experiment}

In this section, the realized experiment is presented. Moreover, the used hardware is introduced and detailed to set the scenario clearly from the beginning. It starts by describing the targeted model and the used dataset. Finally, the experiment to test the presented evasion attack detector is detailed.

\subsection{Scenario}\label{subsec:scenario}

This work follows the scenario presented by Echeberria et al. \cite{echeberria2022understanding}. Therefore, on the one hand, the target model is a deep learning model formed by the VGG16 neural network (feature-extractor) and a dense neural network (classifier). In this case, the classifier has been implemented with sigmoid activation function while \cite{echeberria2022understanding} implemented it with relu activation function. On the other hand, the used dataset is also the breast cancer dataset\footnote{https://www.kaggle.com/paultimothymooney/breast-histopathology-images}, containing two classes of images: non-cancer images (class 0) and cancer images (class 1)  \cite{Janowczyk2016,cruz2014}. This dataset is divided into two subsets: training data (70\%) and test data (30\%), as is usual in this type of study. A stratified train–test split is used to preserve proportions of classes from the original dataset. 

Several adversarial examples are computed using the deep learning model and the test data. For this purpose, three algorithms are used
\begin{itemize}
    \item Fast Gradient Sing Method (FGSM) \cite{DBLP:journals/corr/abs-1811-06492}
    \item Basic Iterative Method (BIM) \cite{Kurakin2019AdversarialWorld}
    \item Projected Gradient Descent Method (PGD) \cite{Madry2018TowardsAttacks}
\end{itemize}
obtaining many corrupted images from each of them. Hence, the new test data will be composed of those corrupted test images and their respective original test images.

\subsection{Experimental process}\label{subsec:experiment}

Therefore, four new datasets are generated by taking the test data of the presented scenario and computing several adversarial test images. The first one contains the $9,489$ adversarial images obtained by the FGSM algorithm and the $9,489$ respective original images from the test data. Hence, the first dataset contains $18,978$ images obtained via original test data. The second one contains $9,378$ adversarial images obtained by the BIM algorithm and the $9,378$ respective original images from the test data. Therefore, the second dataset contains $18.756$ images obtained via original test data. The third one contains $8,087$ adversarial images obtained by the PGD algorithm and the $8,087$ respective original images from the test data. Hence, the third dataset contains $16,174$ images obtained via original test data. Finally, the last dataset is the combination of the all previous data. 

Each dataset is preprocessed as explained in Subsection \ref{subsec:preprocess}, obtaining the three introduced sets ($\Delta(\mathcal{X})^{0.5, 10, 2}_{\phi_{C}}$ \ref{eq:set1}, $Ad(\mathcal{X})_{\phi_{C}}$ \ref{eq:set2} and $\Lambda(\mathcal{X})_{\phi_{C}}$ \ref{eq:set3}) per dataset. Note that in this case, the dataset presented in the scenario is formed by two image classes, which is why the $t$ parameter is taken as $2$. Moreover, the parameter $p$ of the attribute influence is fixed as $0.5$ and the $k$ parameter of the attribute specialization is taken as $10$. Those sets are the data the detector will use to train and learn to detect a type of adversarial example. 

The newly created node attributes are analyzed to observe their scales and to claim that they are not correlated. In fact, it is essential to study if the features are directly correlated with the assigned label, because in case of a correlation occurs, the detector is not necessary.

On the one hand, the statistics of the attributes are computed: mean, std, min, max, $25$ percentile, $50$ percentile, and $75$ percentile. This analysis gives information about the scale of the attributes and shows if any feature out-scale the others. In Table \ref{tab:statistics}, it can be observed that all the attributes are rounding the same values; thus, none will stand out when working with the detector. 

\begin{table}[!ht]
\centering
\caption{The statistics of the generated dataset}
\label{tab:statistics}
\resizebox{0.48\textwidth}{!}{
\begin{tabular}{llllll}
\hline\noalign{\smallskip}
Statistics & Impact & Influence & Input proportion & Special. $0$ & Special. $1$\\
\noalign{\smallskip}\hline\noalign{\smallskip}
mean & -6.4138e-12 & 7.4149e-02 & 4.4245e-02 & 1.0628e-02 & 1.1349e-02\\
std & 3.4789e-02 & 2.6201e-01 & 8.9744e-02 & 5.5708e-02 & 6.8235e-02\\
min & -9.9837e-01 & 0 & 0 & 0 & 0\\
25\% & 0 & 0 & 0 & 0 & 0\\
50\% & 0 & 0 & 0 & 0 & 0\\
75\% & 0 & 0 & 9.9609e-02 & 1.2622e-03 & 6.3471e-04\\
max & 4.9837e-01 & 1 & 1 & 1 & 1\\
\noalign{\smallskip}\hline
\end{tabular}}
\end{table}

\begin{figure}[htp]
    \centering
    \includegraphics[width=0.46\textwidth]{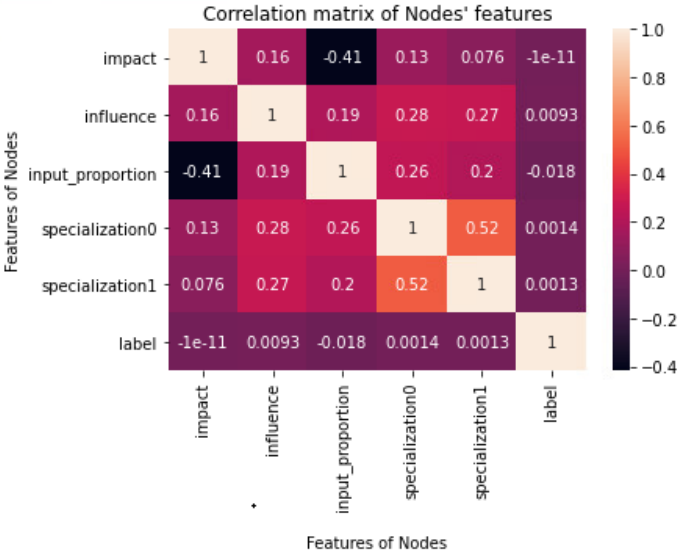}
    \caption{Correlation between the node features in the generated dataset}
    \label{fig:correlation}
\end{figure}

On the other hand, the correlation between the attributes is analyzed. It shows if two characteristics are giving similar or identical information. Moreover, each attribute is compared with the corresponding label to check if any is directly correlated with their label. Figure \ref{fig:correlation} shows that the correlations between the attributes are reasonably low; hence, the attributes give different information. The greatest correlation is between both specializations, but it is expected since there exist neurons that participate in both classes. Moreover, none of them is directly related to the label, so a more complex model is necessary to associate the attributes with the label.  

Therefore, four detectors are developed with the architecture presented in Subsection \ref{subsec:detector}, where each learns to detect one type of presented adversarials. Before the training phase, each dataset is divided into two new sets: training data (80\%) and test data (20\%). 

The training phase starts by assigning each preprocessed dataset to one of the developed detectors. Therefore, each detector is trained with the associated preprocessed dataset, obtaining four trained detectors:
\begin{itemize}
    \item Detector for adversarial images obtained by FGSM.
    \item Detector for adversarial images generated by BIM.
    \item Detector for adversarial images obtained by PGD.
    \item Detector for all adversarial types.
\end{itemize}

\section{Results and Discussion} \label{sec:results}

The results obtained in the experiment presented in Section \ref{sec:experiment} are detailed and discussed in this section. It starts by presenting the motorization of the training phase and the metric obtained by the developed detectors. It follows by comparing obtained results with other similar detectors' results found in the literature.

 \subsection{Results}

\begin{figure}[!ht]
  \centering
  \begin{subfigure}[b]{0.23\textwidth}
    \includegraphics[width=\textwidth]{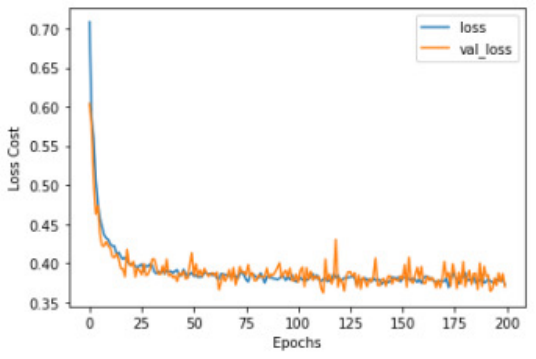}
    \caption{Loss function values}
    \label{fig:fgsm_loss}
  \end{subfigure}
  \begin{subfigure}[b]{0.23\textwidth}
    \includegraphics[width=\textwidth]{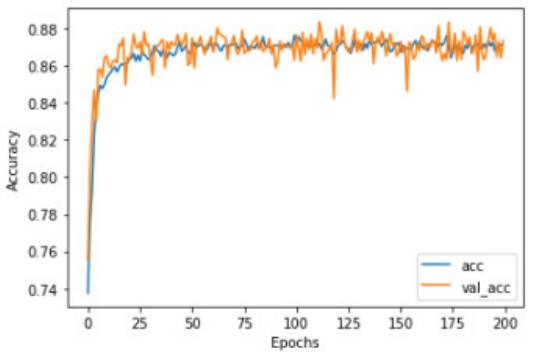}
    \caption{Accuracy}
    \label{fig:fgsm_accuracy}
  \end{subfigure}
  \caption{The monitorization of the loss and accuracy metrics during the training phase of the FGSM detector}
  \label{fig:fgsm_train}
\end{figure}

The first results are associated with the detector that detects the preprocessed data assigned to the FGSM algorithm. In this case, the training data comprises $7,591$ graphs labeled as class $0$ (no adversarial) and $7,591$ class $1$ (adversarial) graphs, while test data is composed of $1,898$ graphs of both classes. Figure \ref{fig:fgsm_train} shows the monitorization of the metric of both training data and test data during the training phase. The loss cost of the test data converges to $0.3623$, obtaining an accuracy of $88\%$ ($0.8832$). In this case, any overfitting is remarkable.

\begin{figure}[!ht]
  \centering
  \begin{subfigure}[b]{0.23\textwidth}
    \includegraphics[width=\textwidth]{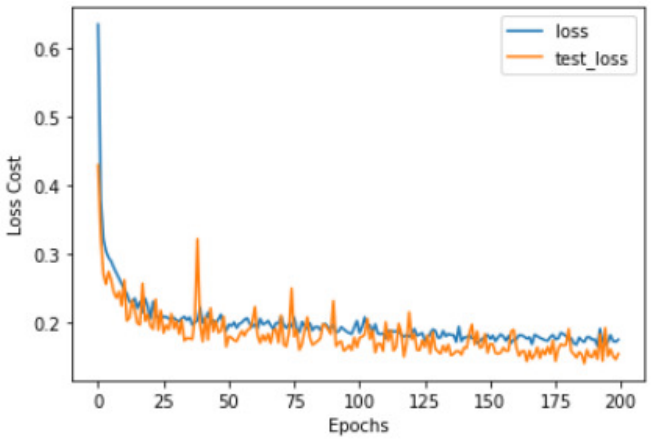}
    \caption{Loss function values}
    \label{fig:bim_loss}
  \end{subfigure}
  \begin{subfigure}[b]{0.23\textwidth}
    \includegraphics[width=\textwidth]{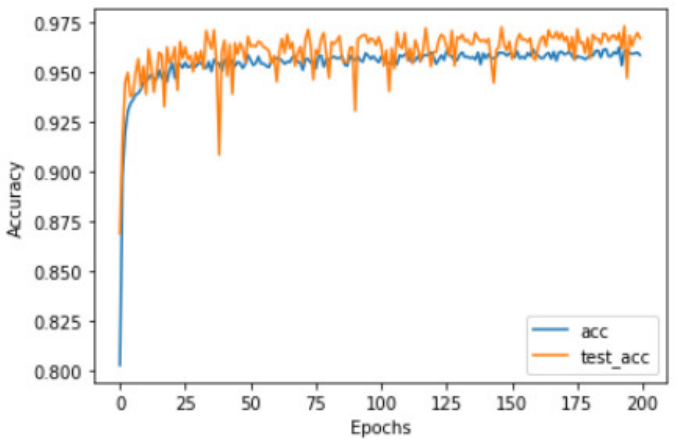}
    \caption{Accuracy}
    \label{fig:bim_accuracy}
  \end{subfigure}
  \caption{The monitorization of the loss and accuracy metrics during the training phase of the BIM detector}
  \label{fig:bim_train}
\end{figure}

The case of the detector that detects the preprocessed data assigned to the BIM algorithm is a bit different. The training data is formed by $7,502$ graphs per class, while test data comprises $1,876$ graphs of both classes. Figure \ref{fig:bim_train} shows the monitorization of the metric of both training data and test data during the training phase. The loss cost of the test data converges to $0.1394$, obtaining an accuracy of $97\%$ ($0.969$). Neither in this case, any overfitting is remarkable.

\begin{figure}[!ht]
  \centering
  \begin{subfigure}[b]{0.23\textwidth}
    \includegraphics[width=\textwidth]{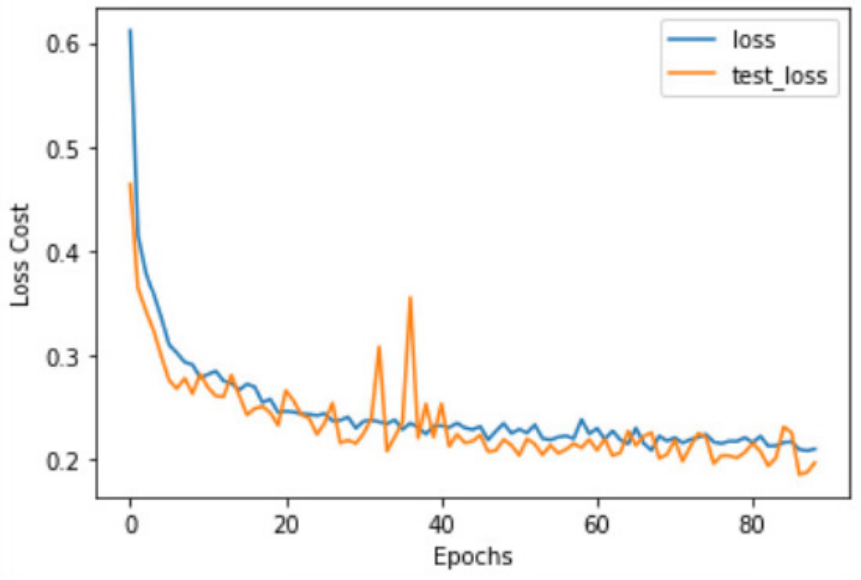}
    \caption{Loss function values}
    \label{fig:pgd_loss}
  \end{subfigure}
  \begin{subfigure}[b]{0.23\textwidth}
    \includegraphics[width=\textwidth]{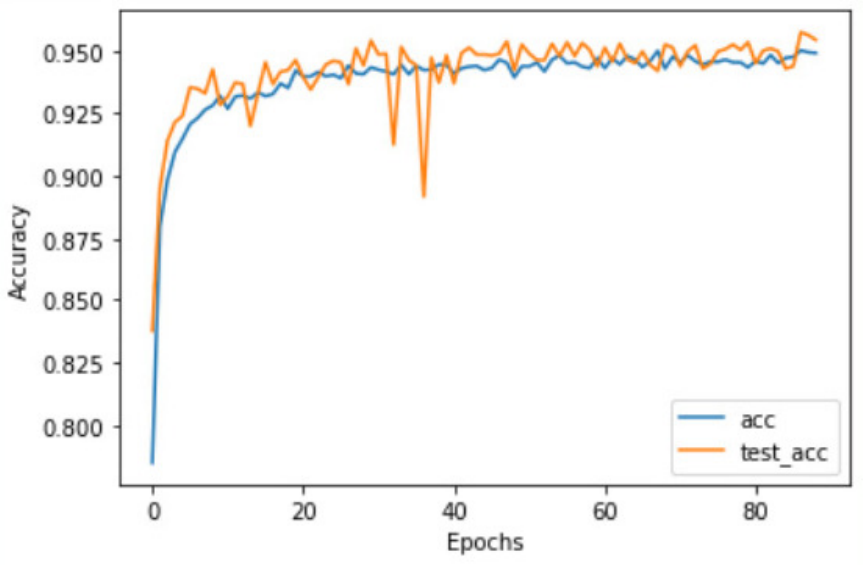}
    \caption{Accuracy}
    \label{fig:pgd_accuracy}
  \end{subfigure}
  \caption{The monitorization of the loss and accuracy metrics during the training phase of the PGD detector}
  \label{fig:pgd_train}
\end{figure}

In the case of the PGD adversarials' detector, the results are the following. On the hand, the training data comprises $6,470$ graphs labeled as class $0$ and $6,469$ class $1$ graphs. On the other hand, the test data is composed of $1,617$ graphs of class $0$ and $1,618$ graphs of class $1$. Figure \ref{fig:pgd_train} shows the monitorization of the metric of both training data and test data during the training phase. The loss cost of the test data converges to $0.1852$, obtaining an accuracy of $95\%$ ($0.9573$). Neither in this case, any overfitting is remarkable.  

\begin{figure}[!ht]
  \centering
  \begin{subfigure}[b]{0.23\textwidth}
    \includegraphics[width=\textwidth]{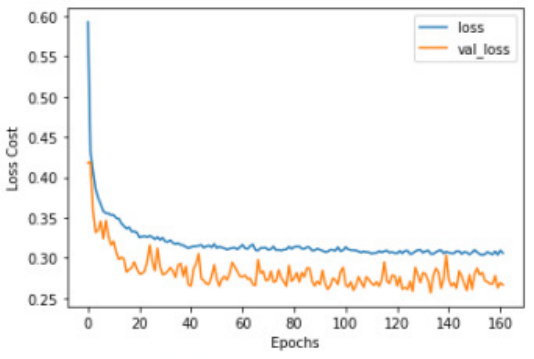}
    \caption{Loss function values}
    \label{fig:total_loss}
  \end{subfigure}
  \begin{subfigure}[b]{0.23\textwidth}
    \includegraphics[width=\textwidth]{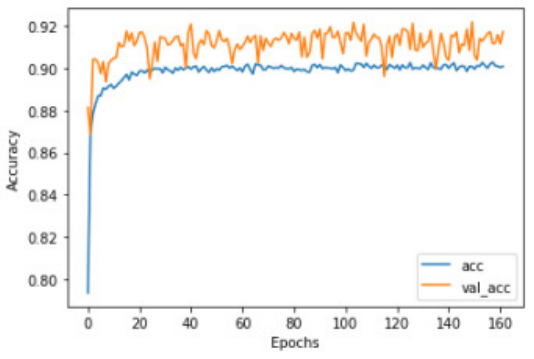}
    \caption{Accuracy}
    \label{fig:total_accuracy}
  \end{subfigure}
  \caption{The monitorization of the loss and accuracy metrics during the training phase of the Total detector}
  \label{fig:total_train}
\end{figure}

Finally, the last detector trained with the preprocessed data obtained from all the algorithms has the following results. The training data comprises $43,126$ graphs of both classes (legitimate or adversarial image), while $10,782$ graphs per class form test data. Figure \ref{fig:total_train} shows the monitorization of the metric of both training data and test data during the training phase. The loss cost of the test data converges to $0.2566$, obtaining an accuracy of $92\%$ ($0.9219$). Neither in this case, any overfitting is remarkable.

Table \ref{tab:results} shows all the accuracies when each detector obtains the lowest loss values during the training phase. Table \ref{tab:results} shows all the accuracies when each detector obtains the lowest loss values during the training phase. It is clear that the BIM attack is the most well-detected with $96.9\%$ of performance, followed by the PGD attack with $95.73\%$ of correct answers. However, the presented detection method reduces the performance in case of an FGSM attack, dropping the accuracy to $88.32\%$. Finally, by combining all the attacks and generating a detector that distinguishes the corrupted images indiscriminately (no matter which algorithm computes it), the total detector is implemented that detects the corrupted images correctly with a $92.19\%$ accuracy.

\begin{table}[!ht]
\centering
\caption{The results obtained from the developed detectors}
\label{tab:results}
\resizebox{0.48\textwidth}{!}{
\begin{tabular}{lllll}
\hline\noalign{\smallskip}
Metrics & FGSM detector & BIM detector & PGD detector & Total detector\\
\noalign{\smallskip}\hline\noalign{\smallskip}
Loss & 0.3623 & 0.1394 & 0.1852 & 0.2566\\
Accuracy & 0.8832 & 0.969 & 0.9573 & 0.9219\\
\noalign{\smallskip}\hline
\end{tabular}}
\end{table}

\subsection{Attribute contribution evaluation}\label{subsec:attribute_evaluation}

In this Subsection, the attributes defined in Section \ref{subsec:attributes} are evaluated separately. Suppose a detector is trained using several features, where each attribute contributes differently to the model's performance. A way to evaluate those contributions is to generate a new detector, training the parameters with a unique attribute for each considered attribute. Now, the computed and the original (all the attributes) performances are compared. This comparison allows deducing the contribution to the detection.

In this case, the presented detector is trained using the dataset introduced in Subsection \ref{subsec:preprocess}. Concretely, Subsection \ref{subsec:experiment} details the parameters of the particular dataset used in our scenario, detailing the parameters. This dataset contains four attributes: impact, influence, input proportion, and specialization. Therefore, per each attribute, a new detector is generated, impact-detector, influence-detector, proportion-detector, and specialization-detector. Each is trained with the data obtained from all the algorithms introduced in Subsection \ref{subsec:scenario}, and only with impact, influence, input proportion, and specialization according to its name.    

\begin{figure}[!ht]
  \centering
  \begin{subfigure}[b]{0.23\textwidth}
    \includegraphics[width=\textwidth]{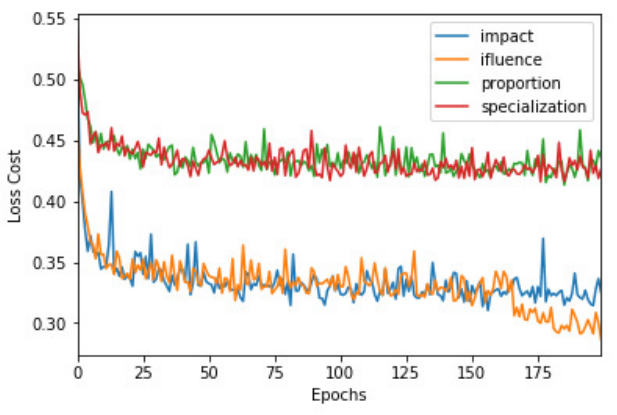}
    \caption{Monitoring of the loss value according to the attribute used in the train}
    \label{fig:attribute_loss}
  \end{subfigure}
  \begin{subfigure}[b]{0.23\textwidth}
    \includegraphics[width=\textwidth]{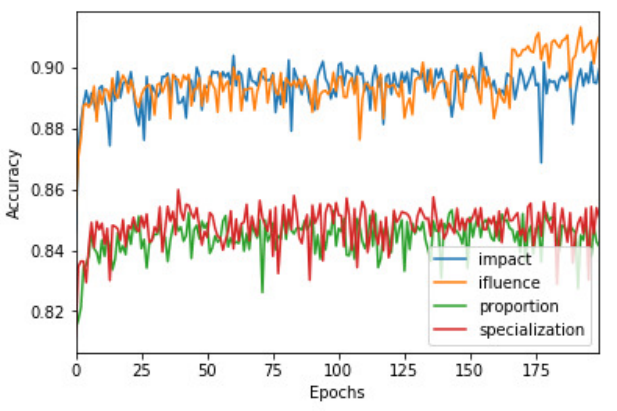}
    \caption{Monitoring of the accuracy value according to the attribute used in the train}
    \label{fig:attribute_acc}
  \end{subfigure}
  \caption{Training phase monitoring for each attribute}
  \label{fig:attribute_monitoring}
\end{figure}

The trained detectors are evaluated, monitoring the loss function of each detector in the training phase. Figure \ref{fig:attribute_monitoring} shows all the loss functions in the training phases. Moreover, Table \ref{tab:attribute_evaluation} details the results that each detector manages. 

\begin{table}[!ht]
\centering
\caption{The results obtained from the attribute detectors}
\label{tab:attribute_evaluation}
\resizebox{0.35\textwidth}{!}{
\begin{tabular}{lll}
\hline\noalign{\smallskip}
Detectors & Accuracy & Loss\\
\noalign{\smallskip}\hline\noalign{\smallskip}
Impact-detector & $0.9047$ & $0.3104$ \\
Influence-detector & $0.9131$ & $0.2865$\\
Proportion-detector & $0.8531$ & $0.4132$\\
Specialization-detector & $0.8598$ & $0.4159$\\
Original detector & $0.9219$ & $0.2566$\\
\noalign{\smallskip}\hline
\end{tabular}}
\end{table}

Table \ref{tab:attribute_evaluation} shows how the impact and influence attributes give more information to the detector to distinguish between original and corrupted images, while the input proportion and specialization attributes obtain worse results when introduced into the detector alone. However, that does not mean they are not helpful in the detection process by combining them with the impact and the influence attributes since they can complement those improving the result. Notice that the evaluation shows that the influence attribute is given the most information among the defined attributes, even though it does not obtain results as good as those obtained by combining all the attributes.

The input proportion attribute has to be mentioned. Remember that it can give more or less information according to the activation function used in the targeted model. Subsection \ref{subsec:attributes} describes that situation, and concretely it details how the sigmoid activation function is one of the possible functions that reduce the information of this attribute. Particularly, the target model in the considered scenario, explained in Subsection \ref{subsec:scenario}, contains the sigmoid activation function in its classifier. Then, the input proportion attribute gives helpful information about the hidden layer of the classifier in this target model, i.e., it provides information about a unique layer of the target model's classifier. Therefore, it is reasonable to think that the input proportion attribute can improve its result if analyzed with another targeted model without a sigmoid function in its classifier.

\subsection{Comparison}

In this Subsection, the results of the presented detector are compared with others found in the literature. It is divided into two Sub-subsections, where the first considers the auxiliary model detectors, and the second takes the detectors based on a similar detection method to be compared.  

\subsubsection{Auxiliary model detectors}

Aldahdooh et al. \cite{aldahdooh2022adversarial} present in 2022 an overview of the evasion detector found in the literature, comparing them with several datasets and giving the performance rate of each one. Moreover, they proportionate the code of those in a GitHub\footnote{https://github.com/aldahdooh/detectors\_review} to run it to obtain the results they mention. This code has been slightly modified to obtain the results according to the scenario defined in this work.

In this case, our proposal is an auxiliary model that receives information about the target model to distinguish between corrupted or not corrupted images. That is why it is compared with the analyzed detectors that consist of an auxiliary model and obtains the best results in the study developed by Aldahdooh et al. \cite{aldahdooh2022adversarial}. Therefore, our detector is compared with LID \cite{ma2018characterizing}, NSS \cite{kherchouche2020detection}, and  KD+BU \cite{feinman2017detecting} in the evasion attacks mentioned in Subsection \ref{subsec:scenario}. 

\begin{table}[!ht]
\centering
\caption{The comparison of the results obtained from the developed detector and the detector of the literature}
\label{tab:comparison}
\resizebox{0.48\textwidth}{!}{
\begin{tabular}{lllll}
\hline\noalign{\smallskip}
Evasion Attacks & LID detector & NSS detector & KD+BU detector & Our detector\\
\noalign{\smallskip}\hline\noalign{\smallskip}
FGSM & $0.6765$ & $0.8076$ & $0.7277$ & $0.8832$\\
PGD & $0.6858$ & $0.8142$ & $0.7701$ & $0.9573$\\
BIM & $0.6830$ & $0.8028$ & $0.7949$ & $0.969$\\
\noalign{\smallskip}\hline
\end{tabular}}
\end{table}

Table \ref{tab:comparison} shows that our detector improves the detection rate in every evasion attack method in the proposed scenario. Our detector obtains the highest results in the BIM attack with $96\%$ detection rate and comparing with the detector that obtains the second best results, NSS, our detector improves the performance by $8\%$ in the FGSM attack method, $14\%$ in the PGD attack method, and $16\%$ in the BIM attack method.      

\subsubsection{Similar method}

This sub-subsection presents a theoretical comparison to show the differences between our detector and a detector developed by Pawlicki et al. \cite{PAWLICKI2020148}, which is based on the same idea as our work. However, as the scenarios implemented are different, the results are difficult to compare, and therefore there will not be an empirical comparison between them.       

Pawlicki et al. \cite{PAWLICKI2020148} presented in 2020 a evasion attack detector based on the analysis of the activation values generated by the targeted model. They mention the novelty of the method, using the activation values to detect if input data is adversarial or not and developing the first detector of this type on the field. It receives the activation values as a long one-dimensional vector, without topological consideration, and considers all the activation of the targeted model.
 
Our work takes this idea to develop the presented detection method. However, this novel detector method is more scalable in the sense of the activation values that the target model generates per received sample. Nowadays, the implemented deep learning models are deeper and deeper, i.e., they contain more and more activation values to analyze. Therefore, this approach will allow the implementation of the detector in this type of model.

Otherwise, our approach considers the topology of the target model as an essential feature for the detection. Moreover, it combines the activation values with that topology, generating new features.

\section{Conclusions}\label{sec:conclusion}

This work presents a novel evasion attack detector that gives good results in the implemented scenario compared with other detectors in the literature. The main novelty of our method is that it incorporates topological information in the detection process. The model is understood as a graph, and its classifier is studied. The graph convolutional network is the ideal technology to take into account the topology of a graph, and therefore it is the base technology of the model detector. The topology is taken into account because the literature demonstrates that it contains essential information to distinguish the model's behaviors. Likewise, this work shows that this characteristic is a feature to consider since the presented detector manages outstanding results.

Otherwise, using the topological information, several attributes are computed. Those attributes give extra information to the detector about the behavior of the targeted model. The developed experiment considers each presented attribute as a unique feature to detect the evasion attack. That is, a performance detecting the evasion attack is obtained for each attribute. Those performances indicate the contribution of each of them in the initial detection process. Therefore, this work demonstrates how much the mentioned attributes are helping in the evasion attack detection, being the influence attribute the one with more helpful information for the detector. At the same time, the input proportion is the less valuable attribute in the detection. However, that result could carry wrong assumptions. As Section \ref{subsec:attribute_evaluation} mentions, the value of the input proportion attribute depends on the activation function that forms the target model's classifier. Indeed, the proposed experiment considers a particular scenario where it takes mainly an activation function that does not give information to that attribute. In other scenarios, the input proportion attribute can gain importance in the detection process.                          

In the future, this detector can be improved by testing different architectures of the presented graph convolutional network. Moreover, more scenarios and experiments should be planned to evaluate this new method to understand it deeply. Even the used hyperparameters can not be optimum, and some exhaustive analysis may optimize them. Otherwise, this detector could be helpful in understanding the relationship between the topological information and the behavior of the neurons through the embedding generated by the model. Concretely, once the detector is trained, many interpretability and explainability techniques exist to show where the model is focusing in the targeted model to detect the evasion attack. That can be considered a method that gives the vulnerable neurons in the model, allowing the generation of local defenses focused on those neurons. Even this defense method can be implemented in detecting other threats that generate anomalies in the activation values, allowing the generation of a defense that detects several attacks apart from the evasion attack, such as poisoning and trojaning attacks. 

\section{Acknowledgements}
This work is funded under the SPARTA project, which has received funding from the European Union Horizon 2020 research and innovation programme under grant agreement No 830892.


\bibliographystyle{elsarticle-num} 
\bibliography{manuscript.bib}





\end{document}